\documentclass[10pt,journal,compsoc]{IEEEtran}
\usepackage{amsmath,amsfonts}
\usepackage{algorithmic}
\usepackage{algorithm}
\usepackage{array}
\usepackage[caption=false,font=normalsize,labelfont=sf,textfont=sf]{subfig}
\usepackage{textcomp}
\usepackage{stfloats}
\usepackage{url}
\usepackage{verbatim}
\usepackage{graphicx} 
\usepackage[nocompress]{cite}
\usepackage{amssymb}
\usepackage{booktabs}
\usepackage{color}
\usepackage{colortbl}
\usepackage{bbm}
\usepackage{multirow}
\definecolor{origin}{rgb}{0.2,0.3,0.9}
\definecolor{mygray}{gray}{.9}
\definecolor{mygray2}{gray}{.8}
\usepackage{hyperref}
\usepackage{url}
\usepackage{graphicx}
\usepackage{amsmath}
\usepackage{mathrsfs}
\usepackage{wrapfig}
\usepackage{booktabs}       
\usepackage{amsfonts}       
\usepackage{nicefrac}       
\usepackage{microtype}      
\usepackage{multirow}
\usepackage{multicol}
\usepackage{makecell}
\usepackage{mathtools}
\usepackage{xspace}
\usepackage{amssymb}
\usepackage{algorithm}
\usepackage{algorithmic}
\usepackage{colortbl}
\usepackage{arydshln}
\usepackage{float}
\usepackage[dvipsnames]{xcolor}
\usepackage{marvosym}
\usepackage{comment}

\definecolor{citecolor}{HTML}{0071BC}
\definecolor{linkcolor}{HTML}{ED1C24}
\definecolor{grey}{HTML}{999999}
\definecolor{green}{HTML}{ABD1BC}
\definecolor{lightblue}{HTML}{B0C4DE}
\definecolor{purple}{HTML}{E3BBED}
\definecolor{orange}{HTML}{ffdab9}

\definecolor{lightmauve}{rgb}{0.86, 0.82, 1.0}
\hypersetup{
    colorlinks=true,
    citecolor=citecolor,
    urlcolor=citecolor,
    linkcolor=black
}

\definecolor{myblue}{RGB}{23,183,241}
\definecolor{mygray}{RGB}{230,230,230}
\newcommand{\hfrcell}[1]{\textcolor{black!40}{#1}}

\begin{document}

\flushbottom
\title{LiFR v2: Completion-Augmented Event Propagation for High-Frequency Dense Prediction}

\author{Tao~Wan$^{*}$, Xiaoshan~Wu$^{*,\dagger}$, Yifei~Yu, Bo~Wang,
Xiaoyang~Lyu, Muxin~Liu, Aoxuan~Pan,\\
Zhongrui~Wang$^{\ddagger}$, and Xiaojuan~Qi$^{\ddagger}$%
\thanks{$^{*}$Tao Wan and Xiaoshan Wu contributed equally to this work.
$^{\dagger}$Xiaoshan Wu is the project lead.
$^{\ddagger}$Corresponding authors: Zhongrui Wang and Xiaojuan Qi.}%
\thanks{Tao Wan, Aoxuan Pan, and Zhongrui Wang are with the School of
Microelectronics, Southern University of Science and Technology,
Shenzhen 518055, China
(e-mail: want2025@mail.sustech.edu.cn;
12631534@mail.sustech.edu.cn;
wangzr@sustech.edu.cn).}%
\thanks{Xiaoshan Wu, Yifei Yu, Bo Wang, Xiaoyang Lyu, Muxin Liu, and
Xiaojuan Qi are with the Department of Electrical and Computer
Engineering, The University of Hong Kong, Hong Kong, China
(e-mail: xiaoshan@connect.hku.hk;
yfyu@connect.hku.hk;
u3009760@connect.hku.hk;
shawlyu@connect.hku.hk;
mxliu@connect.hku.hk;
xjqi@eee.hku.hk).}%
}

\maketitle

\begin{abstract} 
High-frequency dense prediction in dynamic scenes is constrained by low RGB frame rates. Event cameras offer temporally dense but spatially sparse measurements, complementing spatially dense RGB observations. Direct RGB--event fusion cannot fully exploit this complementarity, while event-guided propagation cannot recover newly appearing or disoccluded regions lacking valid RGB support. Building on LiFR-Seg, we present \textbf{LiFR v2} for causal
anytime and streaming dense prediction from the latest RGB frame and events available at each query. Its Event-Guided Completion Module (EGCM) supplements propagated features with spatially gated residuals conditioned on offline-distilled, task-relevant event representations. A History Retrieval Module (HRM) retains these completed states for spatiotemporal retrieval at subsequent queries. Task-specific feature interfaces support semantic segmentation, monocular depth estimation, and joint semantic--depth prediction across different backbones while preserving their original decoders. Moreover, we introduce \textbf{SHF-Emerge}, a synthetic benchmark for rapid object motion, emergence, and disocclusion. On SHF-Emerge, LiFR v2 improves segmentation mIoU from
54.28\% to 56.13\% over LiFR-Seg and reduces 1--10~m depth RMSE by 28.5\% relative to the propagation baseline.
Single-task segmentation and depth inference each exceed 100 FPS on an NVIDIA RTX 5090 GPU.
Code is available at \url{https://github.com/TaoWan0610/LiFR-v2}.
\end{abstract}



\section{Introduction}

\begin{figure*}[t]
    \centering
    \includegraphics[width=1\linewidth]{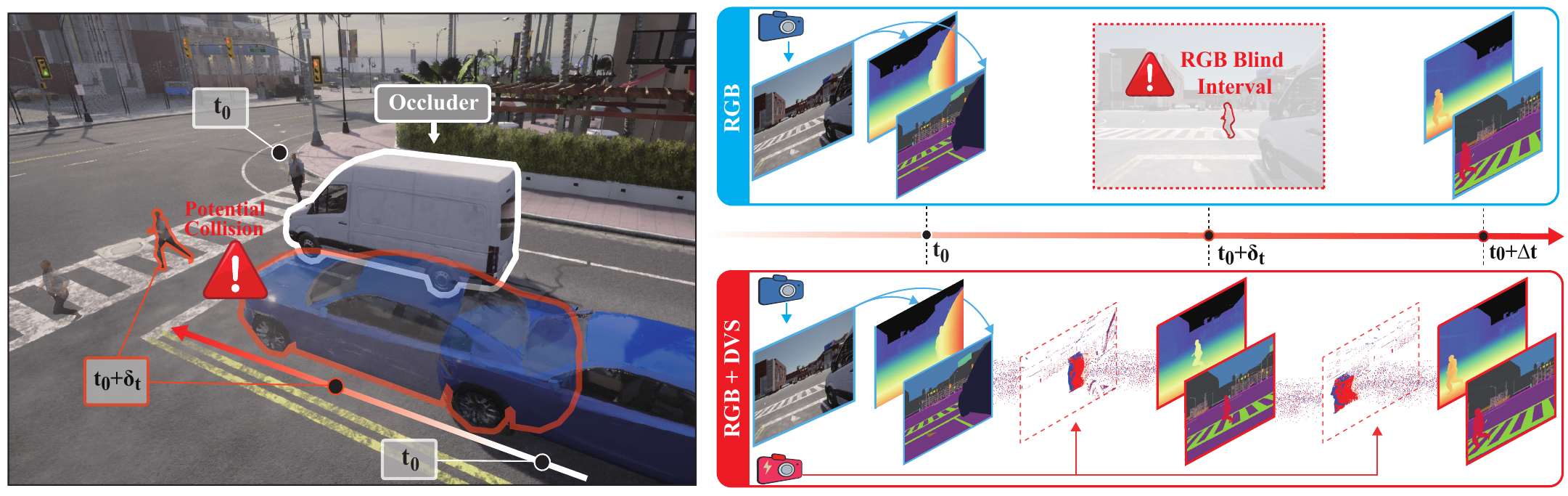}
    \caption{Motivation for causal interframe dense prediction.
A rapidly emerging road user may become visible between
successive RGB observations, leaving a perceptual blind interval
for an RGB-only pipeline. Events recorded after the RGB keyframe
provide evidence of the scene change and support target-time
semantic and depth prediction before the next RGB observation.
The illustration presents the perception scenario rather than
a closed-loop collision-avoidance evaluation.}
    \label{fig:teaser}
\end{figure*}

\IEEEPARstart{D}{ense} prediction provides pixel-wise semantic and geometric understanding for embodied systems such as autonomous vehicles and drones~\cite{cordts2016cityscapes,loquercio2021highspeed,chaplot2020objectgoal,schon2021mgnet,koh2021pathdreamer,liu2026foundationgeo,liu2026optigeo,lyu2026stabilizing}. In dynamic environments, these estimates must remain up to date under rapid viewpoint changes, object motion, and changing visibility~\cite{loquercio2021highspeed,kaufmann2023champion,wu2026eag3r}. However, RGB-based perception is constrained by the finite camera
frame rate, such that no new complete RGB observation is available
between successive frames~\cite{gehrig2024lowlatency}. For an RGB stream at 30~Hz, successive observations arrive 33~ms apart--long enough for an observed object to move substantially or a previously occluded pedestrian to become visible, as illustrated in Fig.~\ref{fig:teaser}. Predictions based on the latest RGB frame can therefore become spatially outdated or miss newly visible content. This motivates \emph{causal high-frequency dense prediction}: estimating the scene state at arbitrary query times using the latest available RGB frame and events observed up to each query, without waiting for the next RGB observation.

Event cameras provide asynchronous measurements of brightness changes with high temporal resolution, offering evidence of scene evolution after the latest RGB observation. Their spatially sparse measurements, however, lack the dense appearance information captured by RGB images. Many RGB--event fusion methods combine these complementary modalities to improve predictions at the RGB frame's timestamp~\cite{zhang2021issafe,zhang2021edcnet,xie2024eisnet,li2025frameevent,gu2026mambaseg,pan2024srfnet,liu2024pcdepth,jing2025unict}. Event-based frame interpolation can reconstruct intermediate images, but methods requiring both endpoint RGB frames cannot provide causal updates before the later frame arrives~\cite{ma2024timelensxl}. 
RAMNet and CFRNet support causal, high-frequency monocular depth estimation through recurrent asynchronous updates and cross-frame-rate collaboration, respectively~\cite{gehrig2021ramnet,liu2025cfrnet}. 
However, their recurrent multimodal updates do not explicitly distinguish between reusing previously observed content and inferring newly visible content. Reliable RGB features can be transported with scene motion, while events provide new evidence to complete content absent from the source frame.

 LiFR-Seg~\cite{wu2026lifrseg} takes a step toward this goal by propagating semantic features from an RGB keyframe to arbitrary query times using event-derived motion and uncertainty-aware splatting. This formulation reuses the image dense predictor's representations while compensating for scene motion using events, achieving segmentation accuracy close to the target-time RGB reference on DSEC without accessing target-time images. Its effectiveness, however, is limited by the source support available
in the RGB keyframe~\cite{niklaus2020softmaxsplatting}. Propagation can
only transport existing features and therefore provides no valid source
features for content that was fully occluded or outside the field of view,
even with accurate motion estimation. Although subsequent events may
reveal such newly visible content, LiFR-Seg does not explicitly convert
this evidence into task-relevant features.

Addressing missing source RGB features for newly appearing objects requires incorporating new scene evidence. We introduce \emph{LiFR v2}, a unified framework for causal anytime and streaming dense prediction that combines feature propagation, event-guided completion, and historical retrieval. The RGB keyframe is encoded once, and its features are propagated using the uncertainty-aware mechanism inherited from LiFR-Seg. An Event-Guided Completion Module (EGCM) then supplements these features with task-relevant event information. For each training dataset, its event encoder
is distilled offline from a frozen, task-trained image encoder using paired
RGB--event training samples, transferring semantic or geometric structure
to the event representation. A learned spatial gate controls where and how strongly event-conditioned corrections are applied, allowing completion to supplement insufficiently supported regions while preserving reliable visual features. We further extend the original temporal memory into a History Retrieval Module (HRM), in which the current propagated representation queries the
RGB anchor and previously completed states through cross-attention augmented with spatial and temporal encodings, retrieving relevant historical context
to refine the current representation and improve temporal consistency
across successive predictions. Completed representations are stored for subsequent queries, allowing content first inferred from events to become part of the retained scene history. The RGB teacher is used only during offline distillation; inference uses the available RGB keyframe and events observed up to the query time.

LiFR v2 operates on intermediate task representations and supports semantic segmentation, monocular depth estimation, and joint semantic--depth prediction through task-specific feature interfaces. These interfaces accommodate different backbone structures while preserving their original task decoders. To evaluate prediction under rapid changes in visibility, we introduce \emph{SHF-Emerge}, a synthetic benchmark featuring object emergence, disocclusion, and large local motion. It emphasizes situations in which newly visible content may lack valid support in the latest RGB observation, providing a focused evaluation of event-guided feature completion.

Experiments on real-world and synthetic benchmarks demonstrate effective causal updates across semantic and geometric prediction tasks. On SHF-Emerge, LiFR v2 improves segmentation from 54.28\% to 56.13\% mIoU over LiFR-Seg. With HRM held fixed, EGCM increases pixel-wise semantic recall
on emerging objects from 71.97\% to 82.31\%, a
10.34-percentage-point gain that supports its effectiveness
in predicting the semantics of newly appearing objects. For monocular depth estimation, LiFR v2 reduces RMSE over the 1--10~m range from 1.564~m to 1.118~m relative to the propagation baseline, a 28.5\% reduction. Streaming evaluation further shows higher segmentation accuracy than LiFR-Seg across successive queries, while single-task segmentation and depth inference each exceed 100 FPS on an NVIDIA RTX 5090 GPU.

\textbf{Extension over the conference version:}
Here, we extend our previous work LiFR-Seg~\cite{wu2026lifrseg} in four aspects.
(i) We introduce event-guided feature completion, combining task-aware event representations with spatially gated corrections to address missing RGB source support.
(ii) We extend temporal memory with spatially and temporally informed retrieval of event-completed states for successive streaming predictions.
(iii) We generalize the framework from semantic segmentation to monocular depth estimation and joint semantic--depth prediction across different backbones.
(iv) We introduce SHF-Emerge and expand the evaluation to examine rapid object emergence, disocclusion, cross-task applicability, and streaming performance.

\looseness=-1

\section{Related Work}
\label{sec:related_work}

\subsection{Event-Based Dense Prediction}
\label{sec:event_representation}

Event-based dense prediction requires transforming asynchronous and
spatially sparse brightness changes into representations that preserve
scene structure and temporal evolution. Early approaches learned
task-specific representations directly from event streams. EV-SegNet
established an early convolutional formulation for semantic
segmentation~\cite{alonso2019evsegnet}, while E2Depth introduced
recurrent processing for monocular depth estimation~\cite{hidalgo2020eventdepth}. Subsequent methods developed more
structured spatio-temporal modeling, including event-prior-guided
attention in EvSegFormer~\cite{jia2023posterior}, recurrent Transformer
modeling in EReFormer~\cite{liu2024ereformer}, hierarchical temporal
memory in HMNet~\cite{hamaguchi2023hmnet}, and edge-semantic and
density-aware modeling in ESEG~\cite{zhao2025eseg}.

A complementary direction transfers supervision from the image domain
to alleviate the scarcity of dense event annotations. EvDistill and
DTL introduce cross-modal knowledge transfer from image-derived
representations~\cite{wang2021evdistill,wang2021dtl}, while ESS aligns
recurrent event and image representations through domain adaptation~\cite{sun2022ess}. More recent approaches exploit large pretrained
models: OpenESS transfers image--text knowledge to event features~\cite{kong2024openess}, whereas Depth AnyEvent and ScaleEvent leverage
visual foundation models for dense event representation learning~\cite{bartolomei2025depthanyevent,chen2026scaleevent}.

Despite these advances, event-only dense prediction remains constrained
by the lack of direct appearance information, particularly in static or
weakly changing regions where event observations are sparse. This
limitation motivates the joint use of conventional images and event
streams for dense prediction.

\subsection{RGB--Event Fusion for Dense Prediction}
\label{sec:rgb_event_fusion}

RGB--event fusion combines dense appearance information from images
with temporally precise event measurements. Early semantic segmentation
methods such as ISSAFE and EDCNet use event-derived dynamics to augment
RGB representations under challenging motion~\cite{zhang2021issafe,zhang2021edcnet}. Later work shifts toward more
adaptive cross-modal interaction: CMNeXt selects useful information
from auxiliary modalities including events~\cite{zhang2023cmnext},
while EISNet explicitly models event activity and modality reliability~\cite{xie2024eisnet}. Recent approaches further accommodate modality
heterogeneity through hybrid ANN--SNN processing~\cite{li2025frameevent} and state-space modeling~\cite{gu2026mambaseg}.

Related developments have emerged for geometric prediction.
Event-Intensity Stereo exploits complementary event and image cues for
dense disparity estimation~\cite{mostafavi2021eventintensity}.
For monocular depth estimation, SRFNet models spatially varying modality
reliability~\cite{pan2024srfnet}, PCDepth learns complementary
high-level patterns~\cite{liu2024pcdepth}, and UniCT Depth combines
local convolutional modeling with global cross-modal interaction~\cite{jing2025unict}. More recent methods introduce iterative depth
hypothesis refinement~\cite{liu2026hypodepth} and asymmetric
state-space representations~\cite{jing2026aimdepth}.

These methods predominantly follow a frame-aligned formulation, in
which an available image and its temporally associated event
representation are jointly processed to improve the corresponding
dense prediction. They therefore focus on multimodal enhancement rather
than updating an earlier RGB representation to an arbitrary interframe
timestamp where no RGB observation is available.
\subsection{Causal Interframe Dense Prediction}
\label{sec:interframe_prediction}
Temporal propagation and memory have been widely studied in frame-based
video perception. Deep Feature Flow and Video Propagation Networks
propagate representations across frames~\cite{zhu2017deepfeatureflow,jampani2017vpn}, while Accel combines
propagated reference features with current-frame features to correct
temporal errors~\cite{jain2019accel}. Memory-based video segmentation
further develops explicit historical retrieval, including
space--time memory matching~\cite{oh2019stm}, improved memory coverage~\cite{cheng2021stcn}, long-term memory organization~\cite{cheng2022xmem}, and streaming memory in SAM 2~\cite{ravi2025sam2}. These methods establish effective mechanisms for
temporal feature propagation and retrieval, but operate on RGB video
streams rather than targeting prediction within intervals lacking a
current RGB observation.

Event cameras enable perceptual updates between successive RGB
observations by providing asynchronous measurements at substantially
higher temporal resolution. In automotive object detection,
event-driven processing has been shown to produce high-frequency predictions
within the RGB interframe interval before the subsequent image becomes
available~\cite{gehrig2024lowlatency}. This formulation, however,
addresses object-level detection rather than pixel-wise dense
prediction. Event-based frame interpolation represents a complementary
strategy for increasing temporal resolution. Methods such as Time Lens
and TimeLens-XL reconstruct intermediate RGB frames from asynchronous
events and two temporally bracketing RGB observations~\cite{tulyakov2021timelens,ma2024timelensxl}. Since the later RGB
observation is required, these interpolation-based formulations are
non-causal for online interframe prediction.

Causal dense prediction instead uses only observations available up to
the query timestamp. RAMNet maintains an asynchronous recurrent
representation jointly updated by frames and events for interframe
monocular depth estimation~\cite{gehrig2021ramnet}, while CFRNet
combines cross-frame-rate multimodal interaction with recurrent temporal
modeling for high-frequency depth prediction~\cite{liu2025cfrnet}. These
methods decouple dense prediction from the native RGB acquisition rate,
but primarily model temporal evolution through recurrent states and
multimodal feature updates.

Our previous work, LiFR-Seg~\cite{wu2026lifrseg}, instead adopts an
explicit propagation formulation for causal interframe semantic
segmentation. Event-derived motion and uncertainty-aware warping
transport RGB-derived semantic features directly to arbitrary
interframe timestamps. However, propagation remains constrained by the
support of the RGB anchor: content absent from the source
representation, such as newly appearing objects and disoccluded
regions, cannot be recovered through feature transport alone.

Building on this formulation, LiFR v2 complements propagation with
event-guided completion for missing source support, reuses completed
representations across streaming queries, and extends causal interframe
prediction beyond semantic segmentation to multiple dense prediction
tasks.

\section{Method}
\label{sec:method}


\subsection{Problem Formulation}
\label{sec:problem_formulation}

Let $\mathbf{I}_{t_0}$ denote an RGB keyframe captured at timestamp
$t_0$, and let the next RGB frame be acquired at $t_0+\Delta t$,
where $\Delta t$ denotes the RGB frame interval. Given an arbitrary
target timestamp
$\tau\in(t_0,t_0+\Delta t]$, our goal is to predict the dense scene
state at $\tau$ without accessing any RGB observation after $t_0$.

We denote the fixed pre-keyframe event context by
$\mathcal{E}^{-}
=\mathcal{E}_{t_0-\Delta t\rightarrow t_0}$,
and the event prefix between the RGB keyframe and the queried timestamp
by
$\mathcal{E}_{\tau}^{+}
=\mathcal{E}_{t_0\rightarrow\tau}$.
Anytime dense prediction is formulated as
\begin{equation}
    \widehat{\mathbf{Y}}_{\tau}
    =
    \mathcal{F}_{\theta}
    \left(
        \mathbf{I}_{t_0},
        \mathcal{E}^{-},
        \mathcal{E}_{\tau}^{+}
    \right),
    \label{eq:anytime_formulation}
\end{equation}
where $\widehat{\mathbf{Y}}_{\tau}$ may represent semantic
segmentation, monocular depth, or multiple dense prediction outputs.
The formulation is strictly causal: neither the target-time RGB frame
$\mathbf{I}_{\tau}$ nor any event occurring after $\tau$ is available
to the model.

LiFR v2 supports both anytime and streaming prediction.
For streaming prediction, we consider an ordered sequence of
target timestamps
$\mathcal{T}=(t_1,\ldots,t_K)$ satisfying
$t_0<t_1<\cdots<t_K\leq t_0+\Delta t$.
At the $k$-th step, the newly arriving event slice
$\Delta\mathcal{E}_k
=\mathcal{E}_{t_{k-1}\rightarrow t_k}$
is appended to the previously observed events, yielding the cumulative
event prefix
$\mathcal{E}_k^{+}
=\mathcal{E}_{t_0\rightarrow t_k}$.
When $K=1$, the model performs anytime prediction. When
$K>1$, it produces a sequence of temporally connected predictions while
retaining historical states across target timestamps.


\subsection{LiFR v2 Framework Overview}
\label{sec:framework}

\begin{figure*}[t]
    \centering
    \includegraphics[width=1\linewidth]{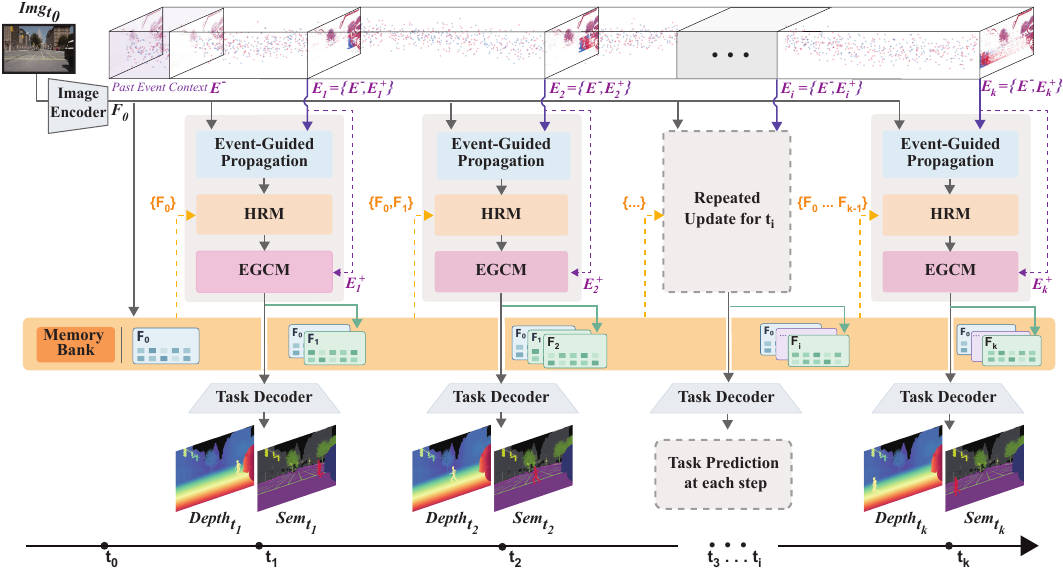}
    \caption{Overall architecture of LiFR v2. The RGB encoder extracts
    anchor features once from the keyframe. Event-derived motion and
    confidence guide uncertainty-aware Softmax Splatting, HRM retrieves and
    refines features using the memory bank, and EGCM uses the current event
    prefix to complete content unavailable through propagation.
    The completed features are decoded and their selected memory-level
    representation is stored for subsequent queries. The illustration shows
    multi-task outputs for semantic segmentation and depth estimation;
    single-task instantiations produce only the corresponding output.
    Memory slots depict the retained feature states, subject to the capacity
    described in Sec.~\ref{sec:hrm}.}
    \label{fig:framework}
\end{figure*}

Fig.~\ref{fig:framework} illustrates the overall architecture of
LiFR v2. The image encoder $\Phi_{\mathrm{img}}$ is executed once on
the RGB keyframe to obtain the anchor representation
\begin{equation}
    \mathbf{F}_0
    =
    \Phi_{\mathrm{img}}
    \left(
        \mathbf{I}_{t_0}
    \right).
    \label{eq:anchor_representation}
\end{equation}
Depending on the task architecture, $\mathbf{F}_0$ may denote either a
single spatial tensor or a multi-scale feature pyramid. The
corresponding memory representation derived from $\mathbf{F}_0$
initializes the historical memory bank $\mathcal{M}_0$.

For the $k$-th prediction at timestamp $t_k$, the shared streaming
update cell is formulated as
\begin{equation}
    \left(
        \widehat{\mathbf{Y}}_{t_k},
        \mathbf{F}_k,
        \mathcal{M}_k
    \right)
    =
    \mathcal{U}_{\theta}
    \left(
        \mathbf{F}_0,
        \mathcal{M}_{k-1},
        \mathcal{E}^{-},
        \mathcal{E}_k^{+}
    \right),
    \label{eq:framework_update}
\end{equation}
where $\mathbf{F}_k$ denotes the completed representation at the
current target timestamp and $\mathcal{M}_k$ is the updated historical
memory.

Each update consists of three stages. First, Event-Guided Propagation
transports the RGB-anchor representation to the queried timestamp
$t_k$ using an event-derived motion field and its estimated reliability.
Second, HRM refines the propagated representation by retrieving relevant
historical context from the memory bank, which is initialized with the
RGB-anchor state and, during streaming inference, augmented with completed
representations from preceding query timestamps. Third, EGCM incorporates
task-relevant event features to complement target-time content that is
inadequately represented by propagation and historical retrieval.

The completed representation $\mathbf{F}_k$ is decoded to produce
$\widehat{\mathbf{Y}}_{t_k}$, and its feature at the selected memory
level is written into the historical bank. Therefore, all target timestamps
share the same RGB anchor and update cell, while their temporal dependency
is maintained through recurrent memory read and write operations.


\subsection{Event-Guided Feature Propagation}
\label{sec:propagation}

We retain the uncertainty-aware feature propagation mechanism of
LiFR-Seg~\cite{wu2026lifrseg} as the temporal propagation backbone of
LiFR v2. For each target timestamp $t_k$, the fixed pre-keyframe event
context $\mathcal{E}^{-}$ and the cumulative target-time event prefix
$\mathcal{E}_k^{+}$ are converted into voxel-grid representations~\cite{zhu2019eventvolume}, 
denoted by $\mathbf{E}^{-}$ and $\mathbf{E}_k^{+}$, respectively.

\begin{figure}[t]
    \centering
    \includegraphics[width=\columnwidth]{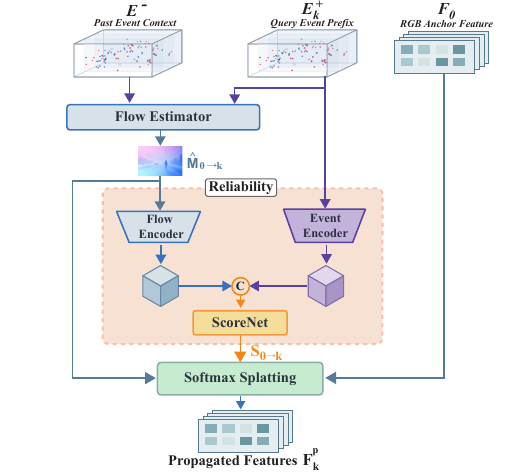}
    \caption{Event-guided uncertainty-aware feature propagation.
    The flow estimator takes the fixed pre-keyframe event voxel
    $\mathbf{E}^{-}$ and cumulative target-time event voxel
    $\mathbf{E}_k^{+}$ to predict the anchor-to-target motion field
    $\widehat{\mathbf{M}}_{0\rightarrow k}$. Encoded flow and current-event
    features are concatenated along channels (C) and passed to ScoreNet
    to estimate the pixel-wise log-precision
    $\mathbf{S}_{0\rightarrow k}$.
    Softmax Splatting uses the motion field and log-precision to propagate
    the RGB-anchor representation $\mathbf{F}_0$ to timestamp $t_k$.}
    \label{fig:propagation}
\end{figure}

As shown in Fig.~\ref{fig:propagation}, E-RAFT~\cite{gehrig2021eraft} 
estimates the anchor-to-target motion field, while ScoreNet
predicts its pixel-wise log-precision:
\begin{equation}
\begin{alignedat}{2}
&\widehat{\mathbf{M}}_{0\rightarrow k}
&{}={}&
\mathcal{F}_{\mathrm{Flow}}
\!\left(
\mathbf{E}^{-},
\mathbf{E}_k^{+}
\right),
\\[2pt]
&\mathbf{S}_{0\rightarrow k}
&{}={}&
\mathcal{F}_{\mathrm{Score}}
\!\left(
\mathbf{E}_k^{+},
\widehat{\mathbf{M}}_{0\rightarrow k}
\right).
\end{alignedat}
\label{eq:motion_and_reliability}
\end{equation}
Here, $\widehat{\mathbf{M}}_{0\rightarrow k}$ denotes the motion field
from $t_0$ to $t_k$, and $\mathbf{S}_{0\rightarrow k}$ measures the
reliability of the estimated motion.

Following LiFR-Seg, uncertainty-aware
Softmax Splatting~\cite{niklaus2020softmaxsplatting} 
propagates the intermediate RGB-anchor representation to the queried
timestamp:
\begin{equation}
    \mathbf{F}_{k}^{\mathrm{p}}
    =
    \frac{
        \overrightarrow{\Sigma}
        \left(
            \exp(\mathbf{S}_{0\rightarrow k})
            \odot \mathbf{F}_0,
            \widehat{\mathbf{M}}_{0\rightarrow k}
        \right)
    }{
        \overrightarrow{\Sigma}
        \left(
            \exp(\mathbf{S}_{0\rightarrow k}),
            \widehat{\mathbf{M}}_{0\rightarrow k}
        \right)
        +\epsilon
    }.
    \label{eq:uncertainty_splatting}
\end{equation}
Here, $\overrightarrow{\Sigma}$ denotes forward splatting according to
the supplied motion field, and the division is element-wise, with the
single-channel weights broadcast across feature channels.
We set $\epsilon=10^{-7}$ to avoid division by zero at target pixels
receiving no source contributions.
The log-precision map controls the contribution of transported features
during forward splatting, thereby reducing the influence of unreliable
motion estimates. The exact feature representation to which
Eq.~\eqref{eq:uncertainty_splatting} is applied depends on the task
architecture and is described in
Sec.~\ref{sec:task_interfaces}.

The propagated representation $\mathbf{F}_{k}^{\mathrm{p}}$ is then
passed to HRM for refinement through historical retrieval.


\subsection{History Retrieval Module}
\label{sec:hrm}

\begin{figure}[t]
    \centering
    \includegraphics[width=\linewidth]{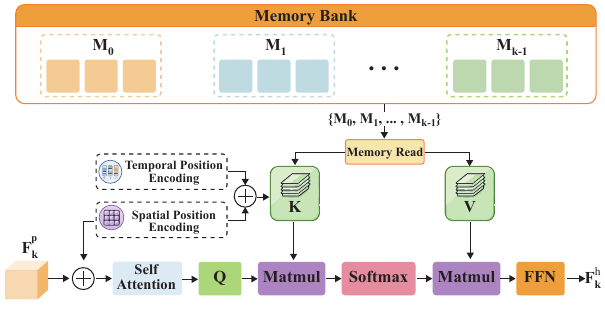}
    \caption{
    Architecture of the History Retrieval Module (HRM).
    The current propagated feature $\mathbf{F}_k^{\mathrm{p}}$ supplies
    the queries, while retained memory features $\mathbf{M}_j$ supply
    the keys and values. The bank is initialized with the RGB-anchor
    feature and subsequently stores completed states from preceding
    queries. Spatial and temporal encodings identify token locations and
    the relative recency of memory states. Historical retrieval and a
    feed-forward network yield the history-enhanced feature
    $\mathbf{F}_k^{\mathrm{h}}$. The schematic omits projection,
    normalization, rotary encoding, residual connections, and block
    repetition for clarity.
    }
    \label{fig:hrm}
\end{figure}

Although LiFR-Seg already employs temporal memory attention on a
high-level semantic representation to improve long-term consistency,
LiFR v2 further augments historical retrieval with explicit spatial and
temporal positional encoding. In addition, the representation completed
by EGCM at each target timestamp is stored in the recurrent memory bank as a historical state and becomes available to subsequent predictions, as illustrated in
Fig.~\ref{fig:hrm}.

The exact memory interface depends on the task backbone. For a
multi-scale backbone, HRM is applied only to a selected deep feature
level, while the remaining feature levels bypass HRM unchanged. For
architectures exposing a single fused temporal representation, HRM is
applied directly to this fused tensor. The task-specific choices are
described in Sec.~\ref{sec:task_interfaces}.

Before predicting at timestamp $t_k$, HRM maintains an ordered
memory bank $\mathcal{M}_{k-1}$ of up to $L$ feature maps.
The initial entry $\mathbf{M}_0$ is derived from the RGB-anchor feature;
subsequent entries $\mathbf{M}_j$ store completed features from earlier
streaming queries. Each retained map is flattened into spatial tokens,
which are concatenated to form $\mathbf{X}_{k-1}^{\mathrm{m}}$.
The memory tokens used to construct the keys are augmented with
spatial and temporal encodings:
\begin{equation}
\widetilde{\mathbf{X}}_{k-1}^{\mathrm{m}}
=
\mathbf{X}_{k-1}^{\mathrm{m}}
+
\mathbf{P}_{k-1}^{\mathrm{m,s}}
+
\mathbf{P}_{k-1}^{\mathrm{m,t}}.
\end{equation}
In this subsection, $\mathbf{F}_k^{\mathrm{p}}$ and
$\mathbf{F}_k^{\mathrm{h}}$ refer to the selected memory feature level.
The propagated feature is flattened and augmented with a scaled spatial
position encoding. Four stacked blocks then apply self-attention,
historical-memory cross-attention, and a feed-forward network~\cite{vaswani2017attention}. 

Let $\mathbf{X}_k^{\mathrm{q}}$ denote the current tokens
after a block's self-attention sublayer. Omitting block and head indices
and projection biases, historical retrieval is expressed as
\begin{equation}
\begin{array}{@{}l@{\;=\;}l@{}}
\mathbf{Q}_k
&
\mathcal{R}_{\mathrm{q}}
\!\left(
\mathbf{X}_k^{\mathrm{q}}\mathbf{W}_Q
\right),
\\[2pt]
\mathbf{K}_{k-1}
&
\mathcal{R}_{\mathrm{m}}
\!\left(
\widetilde{\mathbf{X}}_{k-1}^{\mathrm{m}}\mathbf{W}_K
\right),
\\[2pt]
\mathbf{V}_{k-1}
&
\mathbf{X}_{k-1}^{\mathrm{m}}\mathbf{W}_V,
\\[3pt]
\mathbf{H}_k
&
\operatorname{Softmax}
\!\left(
\frac{
\mathbf{Q}_k\mathbf{K}_{k-1}^{\top}
}{
\sqrt{d_h}
}
\right),
\\[2pt]
\mathbf{A}_k
&
\mathbf{H}_k\mathbf{V}_{k-1}.
\end{array}
\label{eq:hrm_attention}
\end{equation}
Here, $d_h$ is the feature dimension per attention head, and Softmax
normalizes over memory tokens. The projections
$\mathbf{W}_Q$, $\mathbf{W}_K$, and $\mathbf{W}_V$ form queries, keys,
and values, respectively. The spatial encoding
$\mathbf{P}_{k-1}^{\mathrm{m,s}}$ identifies token locations, while the
learned temporal embedding $\mathbf{P}_{k-1}^{\mathrm{m,t}}$ represents
the relative recency of retained memory states.
The operators $\mathcal{R}_{\mathrm{q}}$ and $\mathcal{R}_{\mathrm{m}}$
apply two-dimensional rotary position encoding~\cite{su2024roformer} 
to queries and keys.

The retrieved outputs from all heads are concatenated, projected, and
added to the current tokens through a residual connection.
A feed-forward sublayer updates the tokens through another residual
connection, with pre-normalization used throughout each block.
After the final block, the tokens are normalized and restored to the
spatial layout, yielding $\mathbf{F}_k^{\mathrm{h}}$.
Once EGCM completes the current representation, its feature at the same
selected level is appended as $\mathbf{M}_k$.
The oldest entry is removed when the capacity is exceeded.
HRM thus incorporates historical context at a deep feature level while
leaving the other levels unchanged; EGCM subsequently refines the feature levels selected by the
task-specific completion interface.


\subsection{Event-Guided Completion Module}
\label{sec:egcm}

Event-guided propagation transports information already supported by
the RGB anchor, while HRM retrieves information from the anchor-initialized memory
and, during streaming inference, from completed states of preceding queries. However, neither operation can recover
content that is absent from both the anchor representation and the
historical memory. Such missing-source-support cases arise when an
object newly enters the field of view, an occluded region becomes
visible, or camera motion reveals a previously unseen scene region.

\begin{figure}[t]
    \centering
    \includegraphics[width=\columnwidth]{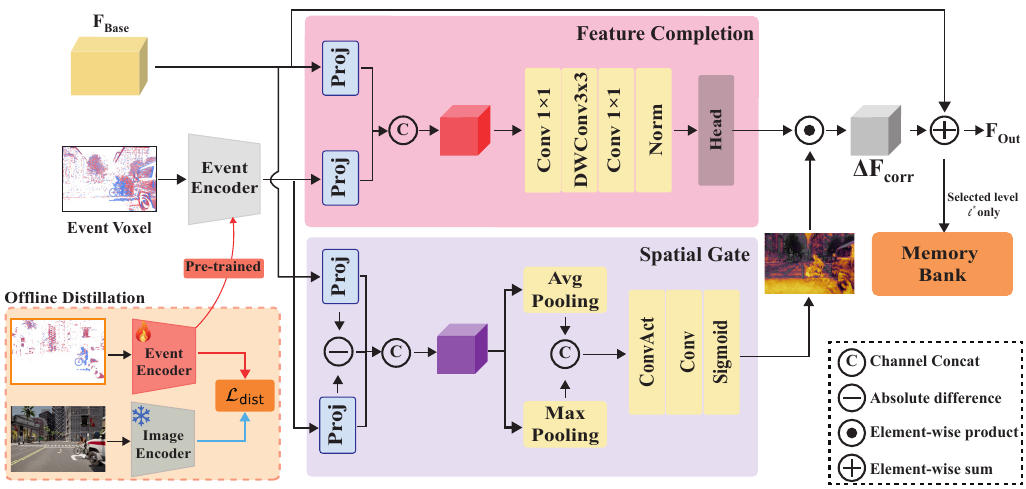}
    \caption{Architecture of the Event-Guided Completion Module
    (EGCM), shown for one feature level. The feature completion module
    combines visual and event features, while the spatial gate also uses
    their absolute difference. Multiplying the residual by the gate gives
    $\Delta\mathbf{F}_{k}^{\mathrm{corr},l}$, which is added to
    $\mathbf{F}_{k}^{\mathrm{base},l}$ to produce
    $\mathbf{F}_{k}^{\mathrm{out},l}$.
    These quantities correspond to $\Delta F_{\mathrm{corr}}$,
    $F_{\mathrm{Base}}$, and $F_{\mathrm{Out}}$ in the diagram.
    During offline distillation, the event encoder receives
    $\mathbf{E}_k^{+}$ and the frozen image teacher receives
    $\mathbf{I}_{t_k}$. Target-time RGB is used only for this
    offline supervision.}
    \label{fig:egcm}
\end{figure}

EGCM addresses this limitation by introducing task-relevant information
from the current event prefix. As shown in
Fig.~\ref{fig:egcm}, EGCM contains an offline-distilled event encoder,
a conditional feature completion module, and a learned spatial gate.

\textbf{Offline task-aware event representation learning.}
We pretrain each event encoder using paired target-interval event voxels
$\mathbf{E}_k^{+}$ and target-time RGB images $\mathbf{I}_{t_k}$.
The event encoder processes $\mathbf{E}_k^{+}$, while a task-trained
image encoder processes $\mathbf{I}_{t_k}$ as a frozen teacher.
Only the event encoder and its feature-alignment layers are optimized
during distillation. Supervision is applied entirely in feature space,
without segmentation-logit or depth-output losses and without
event-activity masking.

Let $\mathbf{S},\mathbf{T}\in\mathbb{R}^{C\times H\times W}$ denote
aligned student and teacher feature maps, with the teacher features
detached from gradient computation. The feature-matching loss is
\begin{equation}
    \mathcal{L}_{\mathrm{feat}}
    =\frac{1}{CHW}\|\mathbf{S}-\mathbf{T}\|_1.
    \label{eq:distillation_feature_loss}
\end{equation}
For relational supervision, the feature maps are converted into token
matrices $\bar{\mathbf{S}},\bar{\mathbf{T}}\in\mathbb{R}^{N\times C}$,
where $N=HW$ and each token is $\ell_2$-normalized along its channel
dimension. We compute the rectified similarity matrices
\begin{equation}
\begin{aligned}
    \mathbf{A}_{TT}&=[\bar{\mathbf{T}}\bar{\mathbf{T}}^{\top}]_+,\\
    \mathbf{A}_{SS}&=[\bar{\mathbf{S}}\bar{\mathbf{S}}^{\top}]_+,\\
    \mathbf{A}_{TS}&=[\bar{\mathbf{T}}\bar{\mathbf{S}}^{\top}]_+,
\end{aligned}
    \label{eq:distillation_relations}
\end{equation}
where $[\cdot]_+$ clips negative entries to zero. A teacher-derived
relation mask 
$\mathbf{R}=\mathbbm{1}[\mathbf{A}_{TT}>\eta]$
, with
$\eta=0.1$, selects positively related token pairs.
Inspired by the relation-based cross-modal distillation strategy in
ScaleEvent~\cite{chen2026scaleevent}, 
self-relation matching transfers the teacher's pairwise feature
structure to the student, while cross-modal relation matching aligns
teacher--student similarities with the same teacher relations:
\begin{equation}
\begin{aligned}
    \mathcal{L}_{\mathrm{self}}
    &=\frac{1}{N^2}
      \|\mathbf{R}\odot(\mathbf{A}_{SS}-\mathbf{A}_{TT})\|_{\mathrm{F}}^2,\\
    \mathcal{L}_{\mathrm{cross}}
    &=\frac{1}{N^2}
      \|\mathbf{R}\odot(\mathbf{A}_{TS}-\mathbf{A}_{TT})\|_{\mathrm{F}}^2.
\end{aligned}
    \label{eq:distillation_relation_losses}
\end{equation}
These losses average over all $N^2$ entries, including masked entries
set to zero. The relation mask depends on teacher feature similarity,
not event activity.
\begin{equation}
\begin{aligned}
\mathcal{L}_{\mathrm{dist}}^{q}
&=\sum_{l\in\mathcal{I}_q}
    \alpha_l^{q}\mathcal{L}_{\mathrm{feat}}^{(l)}\\
&\quad+\sum_{l\in\mathcal{J}_q}\bigl(
    \lambda_{\mathrm{self}}\mathcal{L}_{\mathrm{self}}^{(l)}
    +\lambda_{\mathrm{cross}}\mathcal{L}_{\mathrm{cross}}^{(l)}\bigr),
\end{aligned}
\label{eq:event_distillation}
\end{equation}
where $q\in\{\mathrm{seg},\mathrm{depth},\mathrm{multi}\}$ indexes the
task, and $\mathcal{I}_q$ and $\mathcal{J}_q$ identify the feature-matching
and relation-matching levels, respectively. The coefficients
$\alpha_l^{q}$ weight feature matching, while
$\lambda_{\mathrm{self}}$ and $\lambda_{\mathrm{cross}}$ control the
two relational terms.

For all tasks, we distill task-relevant event representations from
a frozen, task-trained image encoder, while adapting the student event
encoder to the feature organization of the corresponding backbone.

For semantic segmentation, the MiT-B2 encoder of
SegFormer~\cite{xie2021segformer} 
is used as the teacher, with a four-stage MiT-B0 event encoder as the
student. Stage-specific $1\times1$ projections align the student features
with their teacher counterparts. Feature matching is applied at all four
stages, while the self- and cross-relation losses are applied only at
stage 3:
$\mathcal{I}_{\mathrm{seg}}=\{1,2,3,4\}$ and
$\mathcal{J}_{\mathrm{seg}}=\{3\}$.

For depth estimation, the ViT-S encoder of the fine-tuned
MoGe-2~\cite{wang2025moge2} 
model provides the teacher features, and a MobileNet-based event
encoder is used as the student. Intermediate ViT features are fused
into a single spatial tensor. The multi-scale student features are
resized, concatenated, and projected to the same
$C\times H\times W$ representation. The feature, self-relation, and
cross-relation losses are all applied to this fused feature pair.
Denoting the fused level by $\mathrm{f}$, we set
$\mathcal{I}_{\mathrm{depth}}
=\mathcal{J}_{\mathrm{depth}}=\{\mathrm{f}\}$.

For multi-task prediction, the Swin-Tiny encoder~\cite{liu2021swin} 
of the trained MTMamba~\cite{lin2024mtmamba} 
model serves as the teacher, with a four-stage MiT-B0 event encoder
as the student. Channel projection and spatial resizing align the four
student stages with the corresponding teacher features.
As in semantic segmentation,
$\mathcal{I}_{\mathrm{multi}}=\{1,2,3,4\}$ and
$\mathcal{J}_{\mathrm{multi}}=\{3\}$.
Before computing the stage-3 relation matrices, both aligned feature
maps are adaptively average-pooled to
$\min(H_3,32)\times\min(W_3,32)$, while feature matching is computed
on the unpooled features.

After distillation, the learned event encoders and alignment layers
are transferred to the EGCM event branches. We denote the event
features at query time $t_k$ by
$\{\mathbf{Z}_k^{e,l}\}_l$; for depth estimation, this set reduces to
the single fused representation. The target-time RGB teacher branch is
used only during offline distillation and is removed during anytime and
streaming inference.

\textbf{Multi-level conditional completion.}
EGCM operates on a task-specific subset of the feature levels exposed by the corresponding interface. For semantic segmentation, EGCM updates all four pyramid levels
independently. For multi-task prediction, it updates stages 1--3,
while stage 4 bypasses completion. For depth estimation, it operates
on the single fused representation.

At level $l$, let $\mathbf{F}_{k}^{\mathrm{base},l}$ denote the visual
feature provided to EGCM. At the level processed by HRM, this corresponds
to the history-enhanced representation; at the remaining levels, it is
the propagated feature that bypasses HRM. The corresponding event feature
$\mathbf{Z}_{k}^{e,l}$ is spatially aligned with the visual feature.
Modality-specific projections $\Psi_{\mathrm{b},l}$ and
$\Psi_{\mathrm{e},l}$ align their channel dimensions, after which the
two representations are concatenated to predict an additive residual:
\begin{equation}
    \Delta\mathbf{F}_{k}^{l}
    =
    \mathcal{R}_{l}
    \left(
        \Psi_{\mathrm{b},l}
        (\mathbf{F}_{k}^{\mathrm{base},l})
        \Vert
        \Psi_{\mathrm{e},l}
        (\mathbf{Z}_{k}^{e,l})
    \right),
    \label{eq:egcm_residual}
\end{equation}
where $\Vert$ denotes channel concatenation.
The feature completion module consists of a $1\times1$ convolution, a
depthwise $3\times3$ convolution, a second $1\times1$ convolution,
normalization, and an output head. Its output has the same dimensions as
$\mathbf{F}_{k}^{\mathrm{base},l}$ and represents an additive
feature correction.

\textbf{Learned spatial gate.}
Event activity does not necessarily indicate a useful correction at
every spatial location. At each level, EGCM therefore estimates a
single-channel gate
$\mathbf{G}_{k}^{l}\in[0,1]^{1\times H_l\times W_l}$.
The gate branch projects the visual and event features into a shared
embedding, denoted by $\mathbf{u}_{k}^{l}$ and $\mathbf{v}_{k}^{l}$,
respectively. It concatenates
$\mathbf{u}_{k}^{l}\Vert\mathbf{v}_{k}^{l}
\Vert|\mathbf{u}_{k}^{l}-\mathbf{v}_{k}^{l}|$,
so the gate receives both modalities and their element-wise absolute
difference. Average and max pooling along the channel dimension summarize these
concatenated features. Their outputs are concatenated and passed
through convolutional layers and a sigmoid activation.
The gated correction and completed output are
\begin{equation}
\begin{alignedat}{2}
&\Delta\mathbf{F}_{k}^{\mathrm{corr},l}
&{}={}&
\mathbf{G}_{k}^{l}\odot\Delta\mathbf{F}_{k}^{l},
\\[2pt]
&\mathbf{F}_{k}^{\mathrm{out},l}
&{}={}&
\mathbf{F}_{k}^{\mathrm{base},l}
+\Delta\mathbf{F}_{k}^{\mathrm{corr},l}.
\end{alignedat}
\label{eq:egcm_output}
\end{equation}
Here, $\Delta\mathbf{F}_{k}^{l}$ determines the complementary content,
and $\mathbf{G}_{k}^{l}$ controls where and how strongly it is injected.
The gate is broadcast across feature channels during multiplication.

The resulting representation $\mathbf{F}_k$ comprises the
EGCM-updated feature levels together with any bypassed pyramid levels,
or the single completed fused tensor for depth estimation.
It is decoded to obtain $\widehat{\mathbf{Y}}_{t_k}$.
Let $\ell^{\star}$ denote the selected memory feature level. The
completed feature written to the bank and the resulting update are
\begin{equation}
\begin{aligned}
\mathbf{Z}_k^{\mathrm{mem}}
&=\mathbf{F}_k^{\mathrm{out},\ell^{\star}},\\
\mathcal{M}_k
&=\operatorname{Update}
\left(\mathcal{M}_{k-1},\mathbf{Z}_k^{\mathrm{mem}}\right).
\end{aligned}
\label{eq:memory_update}
\end{equation}
For a single-tensor interface, $\ell^{\star}$ selects the sole fused
feature level.
HRM and EGCM therefore serve complementary roles: HRM retrieves
information from previously completed states, whereas EGCM introduces
content newly supported by the current event observations.


\subsection{Task-Specific Feature Interfaces}
\label{sec:task_interfaces}

LiFR v2 is inserted between a task-specific image encoder and its
original decoder through lightweight feature interfaces adapted to the
structure of each backbone. Propagation operates at every feature level exposed by the
corresponding interface, whereas EGCM operates on the task-specific
subset of levels described below. HRM is applied only to a selected
deep level or directly to the fused representation in single-tensor
architectures.

\textbf{Semantic segmentation.}
For SegFormer~\cite{xie2021segformer}, 
the MiT encoder produces a four-stage feature pyramid.
Uncertainty-aware propagation is applied independently to all four
stages. The flow and log-precision maps are resized to each feature
resolution, with the motion-vector magnitudes scaled accordingly.
HRM refines only the stage-3 feature, while the remaining
propagated features bypass the memory module unchanged. EGCM then
updates all four stages, using the history-enhanced representation at
the HRM-selected stage and the propagated representations at the
remaining stages. The completed four-stage pyramid is subsequently
forwarded to the original SegFormer decoder.

\textbf{Monocular depth estimation.}
For MoGe-2~\cite{wang2025moge2}, 
selected intermediate ViT block features are first projected
and fused into a single spatial representation. Propagation, HRM, and
EGCM are then applied sequentially to this fused tensor, immediately
before ConvNeck and the subsequent geometry prediction heads.
Accordingly, LiFR v2 does not independently process either the selected
ViT block features or the five feature levels generated by ConvNeck.
In the ViT-S instantiation, the fused representation is projected to a
384-channel spatial tensor before temporal processing.

\textbf{Multi-task dense prediction.}
For MTMamba~\cite{lin2024mtmamba}, 
the Swin-Tiny encoder~\cite{liu2021swin} 
produces a four-stage feature pyramid.
Uncertainty-aware propagation is applied independently to all four
stages, while HRM refines only the stage-3 feature. EGCM then updates
stages 1--3, using the history-enhanced feature at stage 3 and the
propagated features at stages 1 and 2, while stage 4 bypasses
completion. The resulting pyramid is forwarded to the original
MTMamba decoder.

These task-specific interfaces preserve the original decoder structures
while introducing only lightweight projection and resizing operations
to match feature resolutions and channel dimensions.

\subsection{Training Objective}
\label{sec:training_objective}

After offline event-feature distillation, LiFR v2 is optimized
using only task supervision at the queried timestamp $t_k$.
The distillation losses are not used during subsequent
task-specific training.

The completed representation $\mathbf{F}_k$ is decoded by the
corresponding task decoder to obtain
$\widehat{\mathbf{Y}}_{t_k}$.
The training objective is
\begin{equation}
\mathcal{L}_k
=
\mathcal{L}_{\mathrm{task}}
\left(
\widehat{\mathbf{Y}}_{t_k},
\mathbf{Y}_{t_k}
\right),
\label{eq:training_loss}
\end{equation}
where $\mathcal{L}_{\mathrm{task}}$ follows the objective of the
corresponding dense prediction task.

For semantic segmentation, we use the standard cross-entropy loss.
For monocular depth estimation, we follow the original MoGe-2
training objective, which supervises aligned 3D point predictions
with inverse-depth weighting.
For multi-task prediction with MTMamba, the depth objective
$\mathcal{L}_{\mathrm{depth}}^{\mathrm{multi}}$ is a piecewise-weighted
$\ell_1$ loss over valid pixels, with weights determined by
ground-truth depth intervals. The semantic segmentation and depth
objectives are combined with equal weights:
\begin{equation}
\mathcal{L}_{\mathrm{multi}}
=
\mathcal{L}_{\mathrm{CE}}
+
\mathcal{L}_{\mathrm{depth}}^{\mathrm{multi}}.
\label{eq:multitask_loss}
\end{equation}
The interval boundaries and corresponding weights are provided
in the supplementary material.

\section{Experiments}

\subsection{Experimental Setup}
\label{sec:experimental_setup}

\textbf{Datasets and tasks.}
We evaluate LiFR v2 on six real-world and synthetic RGB--event
benchmarks covering autonomous driving, aerial platforms, and
quadruped robots. DSEC~\cite{gehrig2021dsec} provides synchronized
images and events in real-world urban driving scenes, with semantic
annotations from DSEC-Semantic~\cite{sun2022ess}. M3ED~\cite{chaney2023m3ed}
contains trajectories collected from multiple robotic platforms; we use
its Drone and Quadruped sequences to evaluate aerial and legged-robot
scenarios with substantial ego-motion. SHF-DSEC~\cite{wu2026lifrseg}
is a high-frequency synthetic driving benchmark generated in
CARLA~\cite{dosovitskiy2017carla}. To complement these benchmarks with
targeted evaluations under challenging conditions, we introduce
SHF-Emerge for rapid object emergence, disocclusion, and large local
motion, and use DSEC-Night~\cite{xia2023cmda} to assess robustness
under severe low-light conditions. Detailed construction of SHF-Emerge
is provided in the supplementary material.

\textbf{Evaluation protocols and metrics.}
All causal methods are evaluated using only the RGB keyframe and event
observations available up to the queried timestamp. For semantic
segmentation, we use a prediction horizon of 50\,ms on DSEC, SHF-DSEC,
SHF-Emerge, and DSEC-Night, and 40\,ms on M3ED-Drone and
M3ED-Quadruped. For monocular depth estimation, the prediction horizons
are 100\,ms on DSEC, 80\,ms on M3ED-Drone and M3ED-Quadruped, and
50\,ms on SHF-Emerge. Multi-task prediction follows the same horizons
as monocular depth estimation.

For streaming evaluation, starting from a single RGB keyframe, we query
predictions at 10, 20, 30, 40, and 50\,ms while retaining historical
states across successive queries.

Semantic segmentation is evaluated using mean Intersection-over-Union
(mIoU). For the semantic ablations on SHF-Emerge, we additionally report
pixel-wise semantic recall on emerging objects ($R_{\mathrm{em}}$),
defined as the fraction of valid pixels within their target-time
ground-truth masks that are assigned the correct semantic label.
Counts are pooled across all evaluated masks using the dataset's
semantic taxonomy. Mask selection and construction are detailed in the
supplementary material.

For depth estimation, we report the standard monocular depth metrics~\cite{eigen2014depth}, including absolute relative error (AbsRel),
root mean squared error (RMSE), logarithmic RMSE
($\mathrm{RMSE}_{\log}$), and threshold accuracy $\delta_1$.
Multi-task prediction uses the corresponding semantic and depth metrics.
For SHF-Emerge, we focus on near-field depth because rapid object
emergence and disocclusion are most critical when previously unseen
objects enter the close-range scene, whereas global depth metrics can
be dominated by distant background regions. We therefore evaluate
pixels whose ground-truth depths lie within 1--10\,m and 1--20\,m;
predicted depths are not clipped to these ranges.

\textbf{Architectures and implementation details.}
We use SegFormer-B2~\cite{xie2021segformer} for semantic segmentation,
MoGe-2 ViT-S~\cite{wang2025moge2} for monocular depth estimation, and
MTMamba~\cite{lin2024mtmamba} with a Swin-Tiny backbone~\cite{liu2021swin} for multi-task dense prediction.
The reported MoGe-2 accuracy evaluations use a token budget of 1800.
The event encoder is MiT-B0 for semantic
segmentation and multi-task prediction, and MobileNet-based for depth
estimation. Event streams are represented as voxel grids with 20
temporal bins. For streaming evaluation, HRM maintains up to $L=5$
feature-map states and evicts the oldest entry when the memory is full.

Offline distillation uses task-trained image teachers: SegFormer-B2
for semantic segmentation, MoGe-2 ViT-S for depth estimation, and
MTMamba Swin-Tiny for multi-task prediction.
In Eq.~\eqref{eq:event_distillation}, we set
$\lambda_{\mathrm{self}}=10$ and $\lambda_{\mathrm{cross}}=4$.
The feature-matching weights are
$(\alpha_1^{\mathrm{seg}},\ldots,\alpha_4^{\mathrm{seg}})
=(0.1,0.25,1,1)$,
$\alpha_{\mathrm{f}}^{\mathrm{depth}}=1$, and
$\alpha_l^{\mathrm{multi}}=1/4$ for each multi-task stage.

All models are optimized using AdamW. Detailed task-specific
optimization settings are provided in the supplementary material.
All models are trained on NVIDIA A100 GPUs.

\textbf{Baselines and comparison protocol.}
We compare LiFR v2 against representative paradigms for interframe
dense prediction. We retain the name LiFR-Seg for semantic segmentation and denote its
depth and multi-task adaptations by LiFR.
We first establish two RGB-only references.
The \emph{HFR RGB Reference} applies the task-specific backbone
directly to the RGB observation at the target timestamp
$\mathbf{I}_{t+\delta t}$, while the \emph{LFR RGB Baseline} applies
the same backbone to the reference frame $\mathbf{I}_t$ and evaluates
the prediction against the target-time ground truth.
Together, these two references characterize the effect of the
interframe temporal gap on RGB-based dense prediction.

We next consider interpolation-based and RGB--event fusion paradigms.
Following the interpolation-based comparison in
LiFR-Seg, we use TimeLens-XL (TLX)~\cite{ma2024timelensxl} to reconstruct
the RGB image at the queried timestamp $t+\delta t$ from the two
endpoint RGB frames and the intervening event stream, after which the
corresponding task backbone produces the target-time dense prediction.
Since TLX requires the future endpoint RGB frame, it is treated as a
non-causal interpolation reference.

For RGB--event fusion, we compare EISNet$^{*}$~\cite{xie2024eisnet} and CMNeXt$^{*}$~\cite{zhang2023cmnext} for semantic
segmentation, and RAMNet~\cite{gehrig2021ramnet} and CFRNet~\cite{liu2025cfrnet} for monocular depth estimation.
EISNet and CMNeXt were originally designed for co-temporal RGB--event
fusion, where event observations complement the RGB representation for
segmentation at the corresponding frame. To enable a controlled
comparison under our causal anytime protocol, we adapt them to take the
reference RGB image $\mathbf{I}_t$ together with the target-interval
event prefix $\mathcal{E}_{t\rightarrow t+\delta t}$ and predict the
semantic map at $t+\delta t$, while retaining their original fusion
architectures. RAMNet and CFRNet are evaluated under the same queried
target timestamps for cross-frame-rate depth prediction.

As no directly comparable RGB--event fusion method supports both
semantic segmentation and depth estimation under the same multi-task
setting, we do not include an additional fusion baseline for
multi-task prediction.

Finally, we instantiate the propagation-based formulation of
LiFR-Seg with the corresponding task backbone as a direct baseline
for assessing the additional components introduced in LiFR v2.
All comparisons use the same evaluation splits and query timestamps.
Models requiring task-specific training are trained on the designated
training splits, while DSEC-Night is used only for evaluation. All causal methods are
restricted to observations available no later than the queried
timestamp $t+\delta t$; target- or future-time RGB observations are
used only by the HFR RGB Reference and the interpolation-based
baseline.

\subsection{Anytime Semantic Segmentation}
\label{sec:anytime_segmentation}

\begin{table*}[!t]
\centering
\caption{
Comparison of Anytime Semantic Segmentation on Six Benchmarks.
Results are reported as mIoU (\%).
CS, AT, CM, and EC denote the single-keyframe causal protocol,
anytime prediction, completed-state memory, and event-guided
completion, respectively. $^{*}$ denotes methods adapted to the causal anytime setting. M3ED-D, M3ED-Q, and D-Night denote M3ED-Drone, M3ED-Quadruped, and DSEC-Night, respectively. Best results among causal methods are shown in bold; the HFR RGB Reference is shown in gray.
}
\label{tab:anytime_seg}

\scriptsize
\renewcommand{\arraystretch}{1.15}
\setlength{\tabcolsep}{2pt}

\begin{tabular*}{0.96\textwidth}{@{\extracolsep{\fill}}l*{10}{c}>{\columncolor{gray!10}}c@{}}
\toprule
\textbf{Method}
& \textbf{Input}
& \textbf{CS}
& \textbf{AT}
& \textbf{CM}
& \textbf{EC}
& \textbf{DSEC}
& \textbf{SHF-DSEC}
& \textbf{M3ED-D}
& \textbf{M3ED-Q}
& \textbf{D-Night}
& \cellcolor{gray!18}\textbf{SHF-Emerge} \\
\midrule

\hfrcell{HFR RGB Ref.}
& \hfrcell{$\mathbf{I}_{t+\delta t}$}
& \textcolor{red!60}{$\times$}
& \textcolor{red!60}{$\times$}
& \textcolor{red!60}{$\times$}
& \textcolor{red!60}{$\times$}
& \hfrcell{73.91}
& \hfrcell{65.40}
& \hfrcell{64.57}
& \hfrcell{69.27}
& \hfrcell{41.83}
& \hfrcell{61.18} \\

\midrule

LFR (Baseline)
& $\mathbf{I}_{t}$
& \textcolor{blue}{$\checkmark$}
& \textcolor{red}{$\times$}
& \textcolor{red}{$\times$}
& \textcolor{red}{$\times$}
& 67.67
& 61.73
& 55.23
& 63.20
& 37.44
& 50.03 \\

\midrule
\multicolumn{12}{@{}l}{\textbf{LFR + Interpolation}} \\

TLX + Seg.
& $\mathbf{I}_{t},
 \mathbf{I}_{t+\Delta t},
 \mathcal{E}_{t\rightarrow t+\Delta t}$
& \textcolor{red}{$\times$}
& \textcolor{blue}{$\checkmark$}
& \textcolor{red}{$\times$}
& \textcolor{red}{$\times$}
& 68.17
& 55.89
& 60.60
& 62.92
& --
& 38.91 \\

\midrule
\multicolumn{12}{@{}l}{\textbf{LFR + Fusion}} \\

EISNet$^{*}$
& $\mathbf{I}_{t},
   \mathcal{E}_{t\rightarrow t+\delta t}$
& \textcolor{blue}{$\checkmark$}
& \textcolor{blue}{$\checkmark$}
& \textcolor{red}{$\times$}
& \textcolor{red}{$\times$}
& 68.11
& 61.28
& 58.34
& 62.98
& 37.28
& 54.47 \\

CMNeXt$^{*}$
& $\mathbf{I}_{t},
   \mathcal{E}_{t\rightarrow t+\delta t}$
& \textcolor{blue}{$\checkmark$}
& \textcolor{blue}{$\checkmark$}
& \textcolor{red}{$\times$}
& \textcolor{red}{$\times$}
& 70.13
& 61.40
& 59.56
& 65.52
& 39.38
& 50.05 \\

\midrule
\multicolumn{12}{@{}l}{\textbf{Propagation-based}} \\

LiFR-Seg
& $\mathbf{I}_{t},
   \mathcal{E}_{t-\Delta t\rightarrow t+\delta t}$
& \textcolor{blue}{$\checkmark$}
& \textcolor{blue}{$\checkmark$}
& \textcolor{red}{$\times$}
& \textcolor{red}{$\times$}
& 73.82
& 64.80
& 64.28
& 68.89
& 41.86
& 54.28 \\

\textbf{LiFR v2}
& $\mathbf{I}_{t},
   \mathcal{E}_{t-\Delta t\rightarrow t+\delta t}$
& \textcolor{blue}{$\checkmark$}
& \textcolor{blue}{$\checkmark$}
& \textcolor{blue}{$\checkmark$}
& \textcolor{blue}{$\checkmark$}
& \textbf{74.37}
& \textbf{65.18}
& \textbf{64.99}
& \textbf{70.46}
& \textbf{42.01}
& \textbf{56.13} \\

\bottomrule
\end{tabular*}
\end{table*}

We first evaluate LiFR v2 on anytime semantic segmentation across six
benchmarks. As shown in Table~\ref{tab:anytime_seg}, LiFR v2
substantially outperforms the LFR RGB baseline on all datasets, with
mIoU gains ranging from 3.45 to 9.76 percentage points. In particular,
the improvements reach 9.76 points on M3ED-D, 7.26 points on M3ED-Q,
and 6.10 points on SHF-Emerge. These consistent margins demonstrate
the importance of explicitly updating the stale RGB representation
toward the queried timestamp rather than directly reusing the
keyframe prediction. LiFR v2 also achieves the best performance among
all evaluated causal methods on every benchmark.

Interpolation and direct RGB--event fusion provide partial improvements,
but remain less reliable under large scene changes. TLX performs
competitively on several conventional benchmarks, yet degrades markedly
on SHF-Emerge, where rapid object appearance and disocclusion challenge
image reconstruction. EISNet and CMNeXt benefit from target-interval
events, but their direct fusion formulation does not explicitly transport
the RGB representation toward the queried timestamp. In contrast,
LiFR-Seg establishes a stronger propagation-based baseline by aligning
dense visual features with event-derived motion.

Building on this propagation backbone, LiFR v2 further improves LiFR-Seg
across all six benchmarks. The improvements are modest on DSEC and
SHF-DSEC, while becoming more pronounced on M3ED-Q and SHF-Emerge.
The largest gain is observed on SHF-Emerge, where mIoU increases from
54.28\% to 56.13\%, a 1.85-point improvement over LiFR-Seg. Since this
benchmark contains frequent object emergence, disocclusion, and large
local motion, newly visible regions may lack valid source support in the
RGB anchor. The stronger gain is consistent with the motivation for
event-guided completion, which is designed to supplement propagation
when target-time content lacks valid source support.

\subsection{Generalization to Other Dense Prediction Tasks}
\label{sec:dense_task_generalization}

LiFR v2 operates on intermediate dense representations rather than directly on
task-specific outputs. We therefore investigate whether the proposed temporal
mechanisms are specific to SegFormer-based semantic segmentation or can be
transferred to substantially different encoder--decoder architectures and
dense prediction objectives. We consider monocular depth estimation with
MoGe-2 and multi-task prediction with MTMamba.


\subsubsection{Anytime Monocular Depth Estimation}
\label{sec:anytime_depth}

\begin{table*}[!t]
\centering
\caption{
Anytime monocular depth estimation with MoGe-2 ViT-S. SHF-Emerge is evaluated over the near-field ranges of 1--10\,m and 1--20\,m. For SHF-Emerge, the full metric set is reported over 1--10\,m, while AbsRel and $\delta_1$ are additionally reported over 1--20\,m. Best results among causal methods are shown in bold; ties are highlighted jointly. The HFR RGB Reference is shown in gray.
}
\label{tab:anytime_depth}

\scriptsize
\renewcommand{\arraystretch}{1.15}
\setlength{\tabcolsep}{2.5pt}   

\resizebox{\textwidth}{!}{%
\begin{tabular}{@{}l*{12}{c}*{6}{>{\columncolor{gray!10}}c}@{}}
\toprule
\multirow{2}{*}{\textbf{Method}}
& \multicolumn{4}{c}{\textbf{DSEC}}
& \multicolumn{4}{c}{\textbf{M3ED-D}}
& \multicolumn{4}{c}{\textbf{M3ED-Q}}
& \multicolumn{4}{c}{\cellcolor{gray!18}\textbf{SHF-Emerge (1--10\,m)}}
& \multicolumn{2}{c}{\cellcolor{gray!18}\textbf{SHF-Emerge (1--20\,m)}} \\
\cmidrule(lr){2-5}
\cmidrule(lr){6-9}
\cmidrule(lr){10-13}
\cmidrule(lr){14-17}
\cmidrule(lr){18-19}

& RMSE$\downarrow$
& AbsRel$\downarrow$
& $\delta_1\uparrow$
& $\mathrm{RMSE}_{\log}\downarrow$

& RMSE$\downarrow$
& AbsRel$\downarrow$
& $\delta_1\uparrow$
& $\mathrm{RMSE}_{\log}\downarrow$

& RMSE$\downarrow$
& AbsRel$\downarrow$
& $\delta_1\uparrow$
& $\mathrm{RMSE}_{\log}\downarrow$

& RMSE$\downarrow$
& AbsRel$\downarrow$
& $\delta_1\uparrow$
& $\mathrm{RMSE}_{\log}\downarrow$

& AbsRel$\downarrow$
& $\delta_1\uparrow$ \\
\midrule


\hfrcell{HFR RGB Ref.}
& \hfrcell{3.290} & \hfrcell{0.080} & \hfrcell{0.940} & \hfrcell{0.125}
& \hfrcell{1.846} & \hfrcell{0.094} & \hfrcell{0.928} & \hfrcell{0.112}
& \hfrcell{3.572} & \hfrcell{0.133} & \hfrcell{0.882} & \hfrcell{0.160}
& \hfrcell{1.119} & \hfrcell{0.100} & \hfrcell{0.968} & \hfrcell{0.126}
& \hfrcell{0.112} & \hfrcell{0.950} \\

\midrule

LFR (Baseline)
& 3.960 & 0.104 & 0.913 & 0.170
& 2.761 & 0.132 & 0.874 & 0.156
& 4.375 & 0.140 & 0.840 & 0.224
& 2.729 & 0.181 & 0.936 & 0.231
& 0.191 & 0.921 \\

\midrule
\multicolumn{19}{@{}l}{\textbf{LFR + Interpolation}} \\

TLX + Depth
& 3.368 & 0.085 & 0.934 & 0.130
& 2.607 & 0.141 & 0.843 & 0.153
& 3.670 & 0.141 & 0.859 & 0.167
& 4.171 & 0.407 & 0.604 & 0.366
& 0.411 & 0.595 \\

\midrule
\multicolumn{19}{@{}l}{\textbf{LFR + Fusion}} \\

RAMNet
& 5.377 & 0.155 & 0.766 & 0.213
& 3.884 & 0.144 & 0.812 & 0.191
& 7.335 & 0.150 & 0.796 & 0.263
& 3.467 & 0.458 & 0.510 & 0.534
& 0.458 & 0.489 \\

CFRNet
& 4.049 & 0.102 & 0.894 & 0.151
& 3.657 & 0.129 & 0.838 & 0.180
& 6.060 & 0.168 & 0.754 & 0.260
& 1.733 & 0.121 & 0.930 & 0.196
& 0.139 & 0.909 \\

\midrule
\multicolumn{19}{@{}l}{\textbf{Propagation-based}} \\

LiFR
& 3.381 & \textbf{0.083} & \textbf{0.935} & 0.129
& 1.995 & \textbf{0.093} & \textbf{0.928} & \textbf{0.117}
& 3.741 & 0.141 & 0.850 & 0.180
& 1.564 & 0.148 & 0.909 & 0.174
& 0.130 & 0.943 \\

\textbf{LiFR v2}
& \textbf{3.375}
& \textbf{0.083}
& \textbf{0.935}
& \textbf{0.128}
& \textbf{1.976}
& 0.094
& \textbf{0.928}
& \textbf{0.117}
& \textbf{3.638}
& \textbf{0.123}
& \textbf{0.871}
& \textbf{0.162}
& \textbf{1.118}
& \textbf{0.084}
& \textbf{0.967}
& \textbf{0.126}
& \textbf{0.098}
& \textbf{0.956} \\

\bottomrule
\end{tabular}
}
\end{table*}

Table~\ref{tab:anytime_depth} evaluates LiFR v2 on anytime monocular
depth estimation using MoGe-2 ViT-S. Compared with the LFR RGB baseline,
LiFR v2 substantially reduces the target-time depth error across all
benchmarks. The RMSE decreases by 14.8\% on DSEC, 28.4\% on M3ED-D,
and 16.8\% on M3ED-Q. The improvement is most pronounced on
SHF-Emerge, where the 1--10\,m RMSE drops from 2.729\,m to
1.118\,m, a 59.0\% reduction, while AbsRel decreases from 0.181
to 0.084. These results show that event-driven temporal updating
largely mitigates the degradation caused by relying on a stale RGB
observation.

Compared with interpolation- and fusion-based alternatives, LiFR v2
provides the strongest overall performance among the evaluated causal
methods. TLX and the RGB--event fusion methods improve over the stale
RGB baseline in several settings, but their performance degrades under
strong ego-motion and rapid object emergence. In contrast, explicit
feature propagation provides a stronger target-time representation,
which is further refined by completion and historical retrieval.

Relative to the propagation-based LiFR baseline, LiFR v2 remains
comparable on DSEC and M3ED-D while providing clearer gains on M3ED-Q
and SHF-Emerge. On M3ED-Q, RMSE decreases from 3.741 to 3.638 and
AbsRel from 0.141 to 0.123. On SHF-Emerge, the 1--10\,m RMSE further
decreases from 1.564 to 1.118 and AbsRel from 0.148 to 0.084.
The larger improvements under abrupt visibility changes support the
role of event-guided completion when propagation alone lacks valid
source support.



\subsubsection{Anytime Multi-Task Dense Prediction}
\label{sec:anytime_multitask}

Table~\ref{tab:anytime_multitask} evaluates LiFR v2 with MTMamba
Swin-Tiny for anytime multi-task dense prediction. This setting
provides a more demanding test of architectural generality, since
the temporally updated representation must support semantic and
geometric prediction simultaneously.
\begin{table}[!t]
\centering
\caption{
Anytime multi-task prediction with MTMamba Swin-Tiny.
mIoU is reported in percent. RMSE is reported in meters for DSEC
and M3ED, while AbsRel is reported for SHF-Emerge over the
1--10\,m range.
Best results among causal methods are shown in bold.
}
\label{tab:anytime_multitask}
\scriptsize
\renewcommand{\arraystretch}{1.15}
\setlength{\tabcolsep}{1.8pt}

\resizebox{\columnwidth}{!}{%
\begin{tabular}{l*{6}{c}*{2}{>{\columncolor{gray!10}}c}}
\toprule
\multirow{2}{*}{Method}
& \multicolumn{2}{c}{DSEC}
& \multicolumn{2}{c}{M3ED-D}
& \multicolumn{2}{c}{M3ED-Q}
& \multicolumn{2}{c}{\cellcolor{gray!18}SHF-Emerge} \\
\cmidrule(lr){2-3}
\cmidrule(lr){4-5}
\cmidrule(lr){6-7}
\cmidrule(lr){8-9}

& mIoU$\uparrow$ & RMSE$\downarrow$
& mIoU$\uparrow$ & RMSE$\downarrow$
& mIoU$\uparrow$ & RMSE$\downarrow$
& mIoU$\uparrow$ & AbsRel$\downarrow$ \\
\midrule

\hfrcell{HFR RGB Ref.}
& \hfrcell{71.62} & \hfrcell{4.45}
& \hfrcell{64.26} & \hfrcell{4.58}
& \hfrcell{69.18} & \hfrcell{5.31}
& \hfrcell{59.66} & \hfrcell{0.34} \\

\midrule

LFR (Baseline)
& 55.66 & 4.92
& 49.54 & 4.48
& 59.10  & 5.88
& 50.65 & 0.45 \\

\midrule

\multicolumn{9}{l}{\textbf{LFR + Interpolation}} \\

TLX + Multi-task
& 69.18 & 4.69
& 55.68 & 4.68
& 66.90 & 5.30
& 37.51 & 0.43 \\

\midrule

\multicolumn{9}{l}{\textbf{Propagation-based}} \\

LiFR
& 70.45 & 4.89
& 61.54 & \textbf{4.33}
& 66.18 & 5.48
& 50.28 & 0.40 \\

\textbf{LiFR v2}
& \textbf{70.49} & \textbf{4.84}
& \textbf{61.57} & 4.35
& \textbf{67.25} & \textbf{5.34}
& \textbf{52.57} & \textbf{0.39} \\

\bottomrule
\end{tabular}
}
\end{table}

Compared with the LFR RGB baseline, LiFR v2 yields substantial gains
in semantic segmentation under the shared multi-task setting. The mIoU
improves by 14.83 points on DSEC, 12.03 points on M3ED-D, and
8.15 points on M3ED-Q, with a further 1.92-point gain on SHF-Emerge.
Notably, these semantic improvements are obtained without degrading
the geometric task relative to the LFR baseline: LiFR v2 also reduces
depth error on all four datasets, although the margins are more
moderate than in the single-task depth setting.

Compared with the propagation-based LiFR model, LiFR v2 further
improves semantic mIoU on all four datasets. For depth estimation, it
reduces the error on DSEC, M3ED-Q, and SHF-Emerge, while remaining
comparable on M3ED-D, where RMSE changes only from 4.33 to 4.35.
These results indicate that the proposed temporal update substantially
strengthens semantic prediction while preserving geometric accuracy
within a shared multi-task representation.


\subsection{Streaming Semantic Segmentation}
\label{sec:streaming_segmentation}
We further evaluate LiFR v2 under a stateful streaming setting on
SHF-DSEC. Starting from a single RGB keyframe at $t_0$, newly arriving
events are accumulated in consecutive intervals of 10\,ms, and predictions
are queried at 10, 20, 30, 40, and 50\,ms. Unlike independent anytime
queries, the memory state is retained and updated across successive
predictions.

\begin{figure}[!t]
    \centering
    \includegraphics[width=1\linewidth]{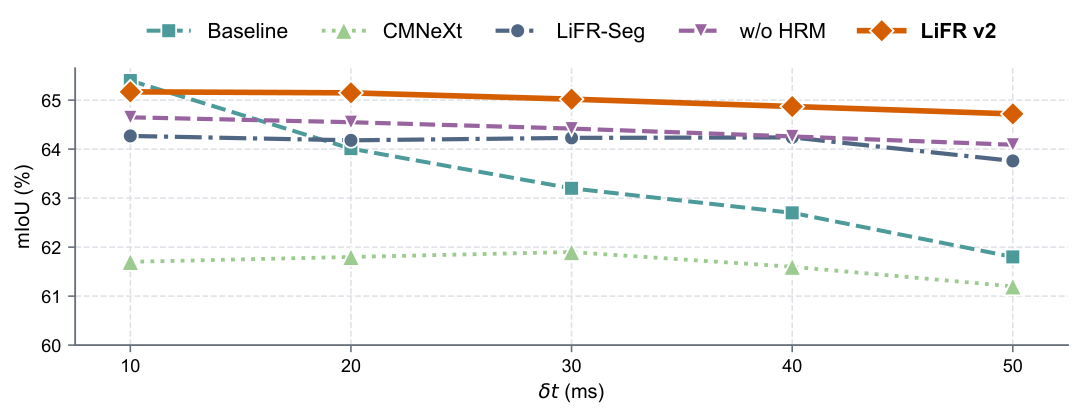}
\caption{
Streaming semantic segmentation on SHF-DSEC from a single RGB
keyframe. Predictions are queried at offsets of 10--50\,ms.
The \emph{w/o HRM} variant retains EGCM but removes historical
retrieval. LiFR v2 consistently outperforms LiFR-Seg, CMNeXt,
and \emph{w/o HRM}, while showing less degradation than the RGB-only baseline.
}
    \label{fig:streaming_hrm_effect}
\end{figure}

As shown in Fig.~\ref{fig:streaming_hrm_effect}, LiFR v2 maintains
more stable accuracy as the query timestamp moves farther from the RGB
keyframe. Although the RGB-only baseline performs slightly better at
10\,ms, its accuracy decreases more rapidly with increasing temporal
offset. LiFR v2 surpasses the RGB-only baseline from 20\,ms onward and
consistently outperforms LiFR-Seg and CMNeXt across the evaluated
streaming timestamps. Moreover, LiFR v2 consistently outperforms the
\emph{w/o HRM} variant, showing that historical retrieval provides
additional benefits beyond event-guided completion alone.

\begin{figure}[!t]
    \centering
    \includegraphics[width=1\linewidth]{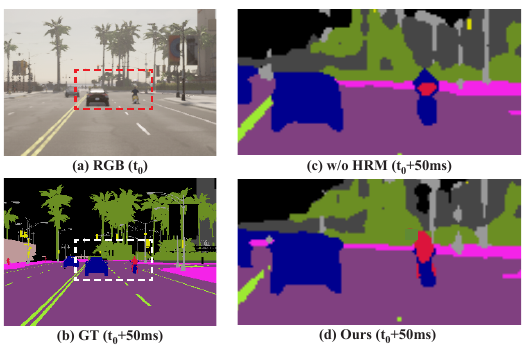}
    \caption{Qualitative effect of historical memory during streaming semantic
segmentation on SHF-DSEC at 50\,ms. (a) Reference RGB frame at $t_0$. (b) Ground truth at
$t_0+50$\,ms. (c) Prediction without HRM. (d) Prediction with HRM.
Dashed boxes highlight regions where historical retrieval improves
semantic consistency.}
    \label{fig:streaming_hrm_comparison}
\end{figure}

The qualitative comparison in
Fig.~\ref{fig:streaming_hrm_comparison} further shows the contribution
of historical retrieval at 50\,ms. Retaining completed states from
preceding queries helps preserve semantic structures that are difficult to maintain without historical retrieval.

\subsection{Computational Cost under Streaming Inference}
\label{sec:computational_complexity}

We report the computational cost and runtime efficiency of the three
LiFR v2 instantiations under streaming inference. After the RGB-keyframe
representation is initialized, it is cached and reused for subsequent
event-driven updates. Parameter counts include all modules active during
inference. GMACs are computed at $640\times440$ resolution and are
amortized over one RGB-keyframe initialization and five consecutive
event-driven updates.

Runtime is measured on a single NVIDIA RTX 5090 GPU with batch size 1.
To reduce Python and kernel-launch overhead during timing, inference is
executed using CUDA Graph replay after warm-up. The reported FPS is
averaged over repeated runs of the same five-query streaming sequence,
including the amortized RGB-keyframe initialization cost.

\begin{table}[!t]
\centering
\caption{
Computational cost of LiFR v2 at $640\times440$ resolution.
The depth model is evaluated with two image-token budgets.
Inference speed is measured on a 5090 GPU.
}
\label{tab:streaming_complexity}

\footnotesize
\renewcommand{\arraystretch}{1.12}
\setlength{\tabcolsep}{3.8pt}

\begin{tabular*}{\columnwidth}{
@{\extracolsep{\fill}}
lrrrr
@{}
}
\toprule
Task
& Tokens
& Params (M)
& GMACs / Output
& FPS $\uparrow$ \\
\midrule

Semantic Seg.
& --
& 60.93
& 96.73
& 108.11 \\

\multirow{2}{*}{Depth}
& 1.2K
& 55.62
& 133.99
& 123.92 \\

& 1.8K
& 55.62
& 184.69
& 106.16 \\

Multi-Task
& --
& 80.82
& 104.42
& 52.63 \\

\bottomrule
\end{tabular*}
\end{table}

As shown in Table~\ref{tab:streaming_complexity}, all LiFR v2
instantiations support high-frequency streaming inference. The semantic
segmentation model operates at 108.11 FPS, while the two depth
configurations achieve 123.92 and 106.16 FPS with 1.2K and 1.8K
image tokens, respectively. The multi-task model jointly predicts
semantic segmentation and depth at 52.63 FPS.

\subsection{Qualitative Analysis}
\label{sec:qualitative_analysis}

\begin{figure*}[!t]
    \centering
    \includegraphics[width=\textwidth,height=0.68\textheight,keepaspectratio]{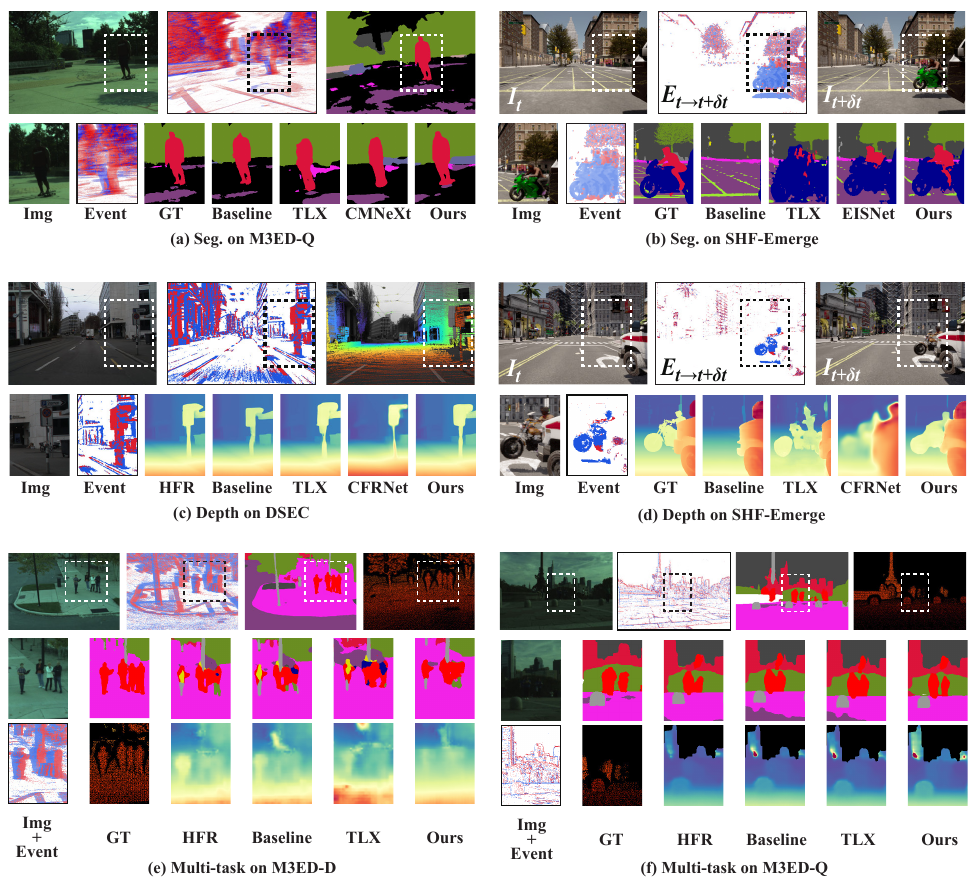}
    \caption{Representative qualitative comparisons for interframe dense prediction.
(a) Semantic segmentation on M3ED-Quadruped with CMNeXt as the RGB--event
fusion baseline. (b) Semantic segmentation on SHF-Emerge with EISNet.
(c) Monocular depth estimation on DSEC and (d) on SHF-Emerge, both
with CFRNet. For (a)--(d), the upper row provides the scene context, including
the reference RGB image and target-interval events. The third panel is
the target-time semantic ground truth in (a), the target-time RGB image
in (b) and (d), and the HFR depth prediction in (c). The lower row shows
enlarged comparisons of the highlighted regions, where the reference map
is GT in (a), (b), and (d), and the HFR depth prediction in (c).
(e) Multi-task prediction on M3ED-Drone and (f) on M3ED-Quadruped,
showing semantic and depth outputs from HFR, LFR, TLX, and LiFR v2.
Target-time RGB images and HFR outputs are shown only as visualization
references and are not used by LiFR v2.}
    \label{fig:qualitative_comparison}
\end{figure*}

Fig.~\ref{fig:qualitative_comparison} presents representative
qualitative comparisons across semantic segmentation, monocular depth
estimation, and multi-task prediction.
Panels (a)--(d) compare single-task outputs within the highlighted
regions, while (e) and (f) show the semantic and depth outputs of the multi-task model.
The SHF-Emerge panels include target-time RGB images to reveal newly
visible content; these images are used only for visualization.
The enlarged DSEC depth reference in (c) is an HFR prediction,
whereas the corresponding reference maps in (a), (b), and (d) are GT.

The LFR baseline remains anchored to the reference RGB observation and
therefore exhibits increasing temporal misalignment at the queried
timestamp. This is visible as displaced object locations, missing newly
visible objects, and inaccurate local geometry. TLX partially compensates
for scene motion by reconstructing the target-time RGB image before task
inference. In the shown high-motion and emergence cases, however,
interpolation artifacts lead to blurred object appearance, indistinct
semantic boundaries, and over-smoothed depth transitions.

For semantic segmentation, the adapted CMNeXt and EISNet baselines
benefit from target-interval event observations and partially recover
moving or newly visible objects. Nevertheless, the displayed examples
still contain fragmented or spatially misaligned contours, particularly
around the pedestrian in (a) and the emerging rider in (b). These
results suggest that direct RGB--event fusion does not fully resolve the
temporal misalignment between the reference RGB representation and the
queried scene state.

For monocular depth estimation, CFRNet captures part of the target-time
scene evolution through cross-frame-rate prediction, but its outputs in
(c) and (d) exhibit smoother object geometry and weaker depth
discontinuities around rapidly moving or newly visible objects.

Across the displayed examples, LiFR v2 yields more coherent target-time
structures, with cleaner semantic contours and sharper local depth
transitions. The difference is particularly evident on SHF-Emerge,
where objects that are absent or heavily occluded in the reference RGB
image become visible at the queried timestamp. In these regions, LiFR v2
recovers more complete semantic and geometric structures than the LFR,
interpolation, and fusion baselines. These qualitative comparisons
further support the benefit of combining explicit feature propagation
with event-guided completion for target-time dense prediction.


\subsection{Ablation Studies}
\label{sec:ablation}

Having evaluated the overall performance and task generality of LiFR v2,
we now examine event-guided completion and task-aware event-feature
distillation through controlled ablations.

\subsubsection{Event-Guided Completion Module}
\label{sec:ablation_egcm}

\textbf{Component ablation.}
We first evaluate the contributions of residual completion and spatial
gating to semantic segmentation.
Table~\ref{tab:egcm_ablation} compares three variants with HRM retained
and the remaining model configurations unchanged.
Ungated Residual adds an event-conditioned residual without spatial
gating, while Full EGCM modulates this residual with a learned
spatial gate.

\begin{table}[!t]
\centering
\caption{
Component ablation of EGCM with HRM retained.
mIoU and pixel-wise semantic recall on emerging objects
($R_{\mathrm{em}}$) are reported in percent.
}
\label{tab:egcm_ablation}

\footnotesize
\renewcommand{\arraystretch}{1.12}
\setlength{\tabcolsep}{2.5pt}

\begin{tabular*}{\columnwidth}{
@{\extracolsep{\fill}}lccc@{}
}
\toprule
Variant
& \makecell[c]{DSEC\\mIoU $\uparrow$}
& \makecell[c]{SHF-Emerge\\mIoU $\uparrow$}
& \makecell[c]{SHF-Emerge\\$R_{\mathrm{em}}\uparrow$} \\
\midrule

(a) No Completion
& 73.80 & 54.28 & 71.97 \\

(b) Ungated Residual
& 74.01 & 55.66 & 81.15 \\

\addlinespace[1pt]
\textbf{(c) Full EGCM}
& \textbf{74.37}
& \textbf{56.13}
& \textbf{82.31} \\

\bottomrule
\end{tabular*}
\end{table}

Ungated Residual improves mIoU by 0.21 percentage points on DSEC
and 1.38 points on SHF-Emerge.
On emerging objects, $R_{\mathrm{em}}$ increases from 71.97\%
to 81.15\%, a gain of 9.18 percentage points.
Adding spatial gating further improves mIoU by 0.36 and 0.47
percentage points on DSEC and SHF-Emerge, respectively, and raises
$R_{\mathrm{em}}$ to 82.31\%.
Residual completion thus provides the larger improvement in
emerging-object recall, with spatial gating yielding an additional gain.

\textbf{Comparison with alternative fusion strategies.}
Table~\ref{tab:completion_fusion} compares Full EGCM with
Naive Addition and Cross-Attention at the completion stage for
semantic segmentation and monocular depth estimation.
Naive Addition directly sums the base representation and
the aligned event-encoder features, whereas Cross-Attention
uses event features as queries and the base representation
as keys and values to retrieve visual information.
Within each task and dataset, all variants use the same frozen,
offline-distilled event encoder and retain a trainable spatial gate.
The propagation, HRM, and decoder configurations, as well as the
training and evaluation protocols, remain unchanged.

\begin{table}[!t]
\centering
\caption{
Completion-stage fusion strategies for separately trained
segmentation and depth models. mIoU and $R_{\mathrm{em}}$ are reported in percent;
$\delta_1$ is reported as a fraction.
SHF-Emerge depth metrics are evaluated over the
1--10\,m ground-truth depth range.
}
\label{tab:completion_fusion}

\footnotesize
\renewcommand{\arraystretch}{1.12}
\setlength{\tabcolsep}{3.0pt}

\begin{tabular*}{\columnwidth}{
@{\extracolsep{\fill}}lccc@{}
}
\toprule
\multicolumn{4}{@{}l}{
\textbf{(a) Semantic Segmentation}
} \\
\addlinespace[2pt]

Variant
& \makecell[c]{DSEC\\mIoU $\uparrow$}
& \makecell[c]{SHF-Emerge\\mIoU $\uparrow$}
& \makecell[c]{SHF-Emerge\\$R_{\mathrm{em}}\uparrow$} \\
\midrule

Naive Addition
& 73.86 & 55.11 & 76.66 \\

Cross-Attention
& 73.97 & 55.76 & 74.74 \\

\addlinespace[1pt]
\textbf{Full EGCM}
& \textbf{74.37}
& \textbf{56.13}
& \textbf{82.31} \\

\bottomrule
\end{tabular*}

\vspace{5pt}

\begin{tabular*}{\columnwidth}{
@{\extracolsep{\fill}}lcccc@{}
}
\toprule
\multicolumn{5}{@{}l}{
\textbf{(b) Depth Estimation}
} \\
\addlinespace[2pt]

\multirow{2}{*}{Variant}
& \multicolumn{2}{c}{DSEC}
& \multicolumn{2}{c}{SHF-Emerge} \\
\cmidrule(lr){2-3}
\cmidrule(lr){4-5}

& AbsRel $\downarrow$
& $\delta_1 \uparrow$
& AbsRel $\downarrow$
& $\delta_1 \uparrow$ \\
\midrule

Naive Addition
& 0.104 & 0.911
& 0.086 & \textbf{0.969} \\

Cross-Attention
& 0.084 & 0.934
& 0.108 & 0.955 \\

\addlinespace[1pt]
\textbf{Full EGCM}
& \textbf{0.083} & \textbf{0.935}
& \textbf{0.084} & 0.967 \\

\bottomrule
\end{tabular*}
\end{table}

For semantic segmentation, Full EGCM achieves the highest mIoU
on both datasets and the highest $R_{\mathrm{em}}$ on SHF-Emerge.
Compared with Naive Addition and Cross-Attention, it improves
$R_{\mathrm{em}}$ by 5.65 and 7.57 percentage points, respectively.
Cross-Attention yields higher overall mIoU than Naive Addition
on SHF-Emerge but lower emerging-object recall, indicating that
improvements in overall segmentation do not necessarily extend
to newly appearing objects.

For depth estimation, Full EGCM achieves the lowest AbsRel
on both DSEC and SHF-Emerge, reaching 0.083 and 0.084,
respectively, compared with 0.104 and 0.086 for Naive Addition,
and 0.084 and 0.108 for Cross-Attention.
It also achieves $\delta_1$ accuracies of 93.50\% on DSEC
and 96.70\% on SHF-Emerge.
These results demonstrate the effectiveness of the proposed
completion mechanism in improving geometric prediction,
particularly under rapid object emergence and disocclusion.

\begin{figure}[!t]
    \centering
    \includegraphics[width=\columnwidth]{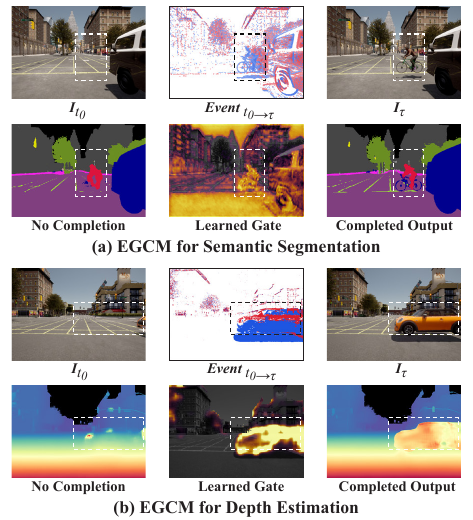}
    \caption{
Qualitative effect of EGCM on (a) semantic segmentation and
(b) monocular depth estimation.
The top row shows the RGB keyframe, target-interval events,
and target-time RGB reference.
The bottom row shows the prediction without completion,
the learned spatial gate, and the completed prediction.
Dashed boxes mark corresponding regions.
    }
    \label{fig:egcm_ablation_visual}
\end{figure}

\textbf{Qualitative results.}
Fig.~\ref{fig:egcm_ablation_visual} illustrates the effect of
completion on semantic and depth predictions.
In the semantic example, the no-completion prediction captures
the rider but misses the bicycle structure.
EGCM recovers a more complete rider--bicycle region, accompanied
by localized spatial-gate responses.
In the depth example, a vehicle absent from the RGB keyframe
becomes visible at the queried timestamp.
The no-completion prediction contains little foreground depth
structure at its location, whereas EGCM produces a distinct
foreground structure with a strong gate response in the
corresponding region.

\subsubsection{Offline Event-Feature Distillation}
\label{sec:ablation_event_distillation}

The event features used by EGCM must encode task-relevant semantic or
geometric structure rather than serving primarily as motion cues for
flow estimation. We therefore compare training from random event-encoder initialization
with training initialized by the offline image-to-event distillation
in Eq.~\eqref{eq:event_distillation}, while keeping all subsequent
training settings unchanged.

\begin{table}[!t]
\centering
\caption{
Effect of offline task-aware event-feature distillation.
For semantic segmentation, we report mIoU and pixel-wise semantic
recall on emerging objects ($R_{\mathrm{em}}$), both in percent;
$R_{\mathrm{em}}$ is evaluated on SHF-Emerge only.
SHF-Emerge depth metrics are evaluated over the
1--10\,m ground-truth depth range.
}
\label{tab:event_distillation}

\footnotesize
\renewcommand{\arraystretch}{1.12}
\setlength{\tabcolsep}{3.0pt}

\begin{tabular*}{\columnwidth}{
@{\extracolsep{\fill}}lccc@{}
}
\toprule
\multicolumn{4}{l}{\textbf{(a) Semantic Segmentation}} \\
\addlinespace[2pt]

Event Encoder
& \makecell[c]{DSEC\\mIoU $\uparrow$}
& \makecell[c]{SHF-Emerge\\mIoU $\uparrow$}
& \makecell[c]{SHF-Emerge\\$R_{\mathrm{em}}\uparrow$} \\
\midrule

w/o Distillation
& 73.95 & 54.91 & 69.78 \\

w/ Distillation
& \textbf{74.37} & \textbf{56.13} & \textbf{82.31} \\

\bottomrule
\end{tabular*}

\vspace{5pt}

\begin{tabular*}{\columnwidth}{
@{\extracolsep{\fill}}
l
cccc
@{}
}
\toprule
\multicolumn{5}{l}{\textbf{(b) Depth Estimation}} \\
\addlinespace[2pt]

Event Encoder
& \makecell[c]{M3ED-D\\AbsRel $\downarrow$}
& \makecell[c]{M3ED-D\\$\delta_1 \uparrow$}
& \makecell[c]{SHF-Emerge\\AbsRel $\downarrow$}
& \makecell[c]{SHF-Emerge\\$\delta_1 \uparrow$} \\
\midrule

w/o Distillation
& \textbf{0.093}
& 0.928
& 0.150
& 0.907 \\

w/ Distillation
& 0.094
& 0.928
& \textbf{0.084}
& \textbf{0.967} \\

\bottomrule
\end{tabular*}
\end{table}

As shown in Table~\ref{tab:event_distillation}, distillation improves
DSEC mIoU from 73.95\% to 74.37\%. On SHF-Emerge, mIoU increases
from 54.91\% to 56.13\%, while $R_{\mathrm{em}}$ rises from 69.78\%
to 82.31\%, a gain of 12.53 percentage points. Without distillation,
$R_{\mathrm{em}}$ is lower than that of No Completion in
Table~\ref{tab:egcm_ablation} (71.97\%), despite higher overall mIoU.
This contrast supports the role of task-aware event pretraining in
improving semantic prediction within emerging-object regions.

A similar pattern is observed for depth estimation. On M3ED-D,
performance remains essentially unchanged, with AbsRel varying from
0.093 to 0.094 and $\delta_1$ remaining at 0.928. In contrast, on
SHF-Emerge over the 1--10\,m range, distillation reduces AbsRel from
0.150 to 0.084 and increases $\delta_1$ from 0.907 to 0.967.
Together, these results indicate that the benefit of task-aware event
distillation becomes more pronounced in scenes involving rapid motion,
object emergence, and abrupt visibility changes.


\section{Conclusion}
We presented \textbf{LiFR v2}, a unified framework for causal anytime and streaming dense prediction using the latest available RGB frame and events observed up to each query. Building on LiFR-Seg, the framework complements event-guided feature propagation with completion and historical retrieval. EGCM combines offline image-to-event distillation with spatially gated corrections to incorporate task-relevant evidence for newly visible content lacking RGB source support. HRM retains and retrieves completed representations, making this newly inferred content available to subsequent predictions. Task-specific feature interfaces extend the framework across semantic and geometric prediction tasks while preserving the original RGB decoders.

Experiments on real-world and synthetic benchmarks demonstrate the framework's applicability to semantic segmentation, monocular depth estimation, and joint semantic--depth prediction across different backbones. SHF-Emerge highlights the benefits of completion for semantic recognition and near-field depth estimation under rapid object emergence and disocclusion. Streaming evaluations show higher segmentation accuracy than LiFR-Seg across the tested query times, while single-task segmentation and depth inference each exceed 100 FPS on an NVIDIA RTX 5090 GPU. These findings show that high-frequency dense perception benefits from both updating existing RGB-derived features and incorporating new event evidence, with completed representations retained to support successive predictions before the next RGB observation arrives.

\section*{Acknowledgments}

This work was supported in part by the Research Grants Council (RGC) 
of Hong Kong under the General Research Fund (Grant Nos. 17202422, 
17212923, and 17215025), the Theme-based Research Scheme 
(Grant No. T45-701/22-R), and the Strategic Topics Grant 
(Grant No. STG3/E-605/25-N), and in part by Voyager Research, 
DiDi Chuxing. Part of this work was conducted in the JC STEM Lab 
of Robotics for Soft Materials, funded by The Hong Kong Jockey Club 
Charities Trust. This work was also partially conducted at the 
AI Chip Center for Emerging Smart Systems (ACCESS), supported by 
the InnoHK initiative of the Innovation and Technology Commission 
of the Hong Kong Special Administrative Region Government.

\bibliographystyle{IEEEtran}
\bibliography{references}

\end{document}